\documentclass{article}
\usepackage{arxiv}
\usepackage[pages=all, color=black, position={current page.south}, placement=bottom, scale=1, opacity=1, vshift=5mm]{background}
\usepackage{amsmath}
\usepackage{amsthm}
\usepackage{amssymb}
\usepackage{graphicx}
\usepackage{scalerel}
\usepackage{tikz}
\usetikzlibrary{intersections}
\usetikzlibrary{spy}
\usetikzlibrary{shapes.geometric}
\usepackage{comment}
\usepackage{booktabs}
\usepackage{amsfonts}
\usepackage{subfig}
\usepackage{epstopdf}
\usepackage{booktabs}
\usepackage{float}
\usepackage{multirow}
\usepackage{tabu}
\usepackage{soul}
\usepackage{tabulary}
\usepackage{enumerate}
\usepackage[nomessages]{fp}
\usepackage{pgfkeys}
\usepackage{adjustbox}
\usepackage{sidecap}
\usepackage{array}
\usepackage{eqparbox}
\usepackage{caption}
\usepackage{tabulary}
\usepackage{filecontents}
\usetikzlibrary{intersections}
\usepackage{mathtools}
\mathchardef\mhyphen="2D % Define a "math hyphen"

\usepackage[utf8]{inputenc}
\usepackage{hyperref}
\hypersetup{
	unicode,
	pdfauthor={Author One, Author Two, Author Three},
	pdftitle={A simple article template},
	pdfsubject={A simple article template},
	pdfkeywords={article, template, simple},
	pdfproducer={LaTeX},
	pdfcreator={pdflatex}
}

\usepackage[sort&compress,numbers,square]{natbib}
\theoremstyle{plain}

\theoremstyle{definition}

\graphicspath{{fig/}}
\newcommand{\ourname}{Deep-PxAF\xspace} 
\usepackage{algorithm, algpseudocode} % use algorithm and algorithmicx for typesetting algorithms
\usepackage{mathrsfs} % for \mathscr command

\usepackage{lipsum}
\usepackage{xcolor}
\usepackage{xspace}
\usepackage{paralist}
\definecolor{carmine}{rgb}{0.59, 0.0, 0.09}
\title{Accurate Detection of Paroxysmal Atrial Fibrillation with
Certified-GAN and Neural Architecture Search}

\date{} 				

\author{{Mehdi~Asadi} \\
	Department of Electrical Engineering, Tarbiat Modares University, Tehran, Iran \\
	\texttt{mehdi.asadi@modares.ac.ir} \\
	\And
	\hspace{1mm}Fatemeh~Poursalim\\
	Shiraz University of Medical Science, Shiraz, Iran\\
	\texttt{fpoursalim72@gmail.com} \\
	\AND
	\hspace{1mm}Mohammad~Loni \\
         School of Innovation, Design and Engineering, M\"alardalen University, V\"aster\aa s, Sweden \\
	\texttt{mohammad.loni@mdu.se}
 \AND
        \hspace{1mm}Masoud~Daneshtalab \\
         School of Innovation, Design and Engineering, M\"alardalen University, V\"aster\aa s, Sweden \\
	\texttt{masoud.daneshtalab@mdu.se}
 \AND
 \hspace{1mm}Mikael~Sj\"odin \\
         School of Innovation, Design and Engineering, M\"alardalen University, V\"aster\aa s, Sweden \\
	\texttt{mikael.sjodin@mdu.se}
 \AND
 \hspace{1mm}Arash~Gharehbaghi \\
         School of Information Technology, Halmstad University, Halmstad, Sweden \\
	\texttt{arash.ghareh-baghi@hh.se}
}

\date{}

\renewcommand{\shorttitle}{Accurate Detection of Paroxysmal Atrial Fibrillation with Certified-GAN and Neural Architecture Search}

\begin{document}
%     \begin{center}
%         \textcopyright \textbf{2022 IEEE. Personal use of this material is permitted.
%             Permission from IEEE must be obtained for all other uses, in any current or future
%             media, including reprinting/republishing this material for advertising or promotional
%             purposes, creating new collective works, for resale or redistribution to servers or
%             lists, or reuse of any copyrighted component of this work in other works.}
%     \end{center}
%   \newpage
\maketitle
\begin{abstract}
	This paper presents a novel machine learning framework for detecting Paroxysmal Atrial Fibrillation (PxAF), a pathological characteristic of Electrocardiogram (ECG) that can lead to fatal conditions such as heart attack.
To enhance the learning process, the framework involves a Generative Adversarial Network (GAN) along with a Neural Architecture Search (NAS) in the data preparation and classifier optimization phases.
The GAN is innovatively invoked to overcome the class imbalance of the training data by producing the synthetic ECG for PxAF class in a certified manner. The effect of the certified GAN is statistically validated.
Instead of using a general-purpose classifier, the NAS automatically designs a highly accurate convolutional neural network architecture customized for the PxAF classification task.
Experimental results show that the accuracy of the proposed framework exhibits a high value of 99\% which not only enhances state-of-the-art by up to 5.1\%, but also improves the classification performance of the two widely-accepted baseline methods, ResNet-18, and Auto-Sklearn, by $2.2\%$ and $6.1\%$. 
\end{abstract}
\keywords{Electrocardiogram (ECG), Paroxysmal Atrial Fibrillation (PAF), Data Augmentation, Neural Architecture Search}
\section{Introduction}
\label{sec:introduction}
Recent progresses in artificial intelligence and Deep Learning (DL) methods created a leap toward automatic decision-making in various domains including health and medicine. Sophisticated Deep Learning (DL) methods have been proposed for classifying biological signals \cite{Gharehbaghi-ITNNLS}, including heart sound \cite{Gharehbaghi2018-VSD, Gharehbaghi-ASC} and electrocardiogram \cite{ebrahimi2020review}. Electrocardiograph (ECG) is a recording of the electrical activity of the heart, reflecting possible disorders in heart function. Paroxysmal Atrial Fibrillation (PxAF) is a disorder in the electrical activity of the heart that can lead to adverse events such as cardiac stroke \cite{Hisashi2018}. Screening patients with PxAF is currently performed by physicians in their clinical practice, and the development of a reliable system for automated detection of PxAF is a need for any healthcare system. Several methods have been proposed for detecting PxAF from the ECG signal \cite{ebrahimi2020review, pourbabaee2018deep, gilon2020forecast, shashikumar2018detection}, from which the DL-based ones are considered as the state-of-the-art of this topic \cite{pourbabaee2018deep, surucu2021convolutional}. Nevertheless, accurate detection of PxAF is still an open research question \cite{pourbabaee2018deep, surucu2021convolutional}).

 We hypothesize two issues that could have led to inaccurate PxAF diagnosis. Firstly, the class imbalance is commonly seen in most of the public ECG databases, where the size of the class with PxAF arrhythmia is by far smaller than the one with normal cases. Secondly, the backbone architectures used in the state-of-the-art studies may not be optimal as they were manually designed for image classification tasks.

One solution to tackle the first issue is to increase the group size of the minority class, i.e., the PxAF class \cite{khushi2021comparative}, by producing synthetic data from the real ones. Patients' real data are being recorded electronically by healthcare providers and private industries. However, the recorded data is hardly accessible to scientists due to patient privacy concerns. Even when researchers are able to access high-quality data, they must ensure that the data is properly used and protected in a legal and ethical manner which is a time-consuming process \cite{goncalves2020generation}.

Generating synthetic medical data has been broadly explored for various sorts of medical data including physiological signals \cite{NPJ2020}. Synthetic ECG data has been reported as the case study in several reports (Section~\ref{sec:related_work:data_augmentation}). Recently, Generative Adversarial Networks (GANs) have demonstrated impressive performance in medical data augmentation. However, the synthetic ECGs, generated by GAN, are mostly immature to be used as the training data due to morphological irrelevance, and thus, leveraging them in the training process can mislead the classifier. As we will see in the sequels, this important point is elaborately considered by the proposed method.

Neural Architecture Search (NAS), as an automated technique for designing artificial neural networks, has recently received attention from researchers and engineers. It provides a solid tool to achieve an optimized architecture for the problem of designing an optimal machine learning solution. Applicability of this technique has been explored in different domains such as biomedical engineering, in which classification of physiological signals is an important challenge \cite{fayyazifar2021accurate, odema2021eexnas, lv2021heart, liu2021automatic}.

In this paper, we propose an original framework for detecting PxAF arrhythmia based on an enhanced combination of GAN and NAS. The framework is composed of three compartments: 1) data enrichment, 2) signal processing, and 3) machine learning compartments. The proposed framework introduces innovative ideas in the methodologies employed for this important research question. It proposes the use of a GAN architecture for data enrichment in a new manner, named certified-GAN, in conjunction with the original signal processing and machine learning methods. The performance of the framework is statistically evaluated both holistically and independently for each compartment. The accuracy of the framework in detecting PxAF was estimated to be 99\%, exhibiting a considerable improvement in the state-of-the-art.

To the best of our knowledge, this paper is the first study proposing an automatic methodology for certified synthetic data generation and designing an accurate CNN architecture for PxAF detection. We name this combination of certified-GAN and NAS for PxAF detection as \ourname. The contributions of this paper are:

\begin{itemize}
    \item A novel data enrichment method is proposed that enables the generation of the certified synthetic PxAF samples based on the recommendations of an expert physician (Section~\ref{sec:method:synthetic_data_generation}).  
    \item A novel data pre-processing approach is proposed to improve the detection performance (Section~\ref{sec:method:signal_processing}).
    \item A cell-based neural architecture search method is employed to design a specialized CNN architecture for the PxAF detection task (Section~\ref{sec:method:CNN_optimization}).
    \item We provide extensive experiments to demonstrate the effectiveness of \ourname (Section~\ref{sec:results}). Plus, we discuss the reproducibility results of the proposed method (Section~\ref{sec:discussion}).
\end{itemize}

Results show that \ourname achieves higher accuracy compared to handcrafted DL architectures and automated machine learning (AutoML) tools on the PhysioNet PxAF database \cite{clifford2017af}. Moreover, \ourname shows stable results with marginal differences with multiple repetitions, confirming the reproducibility of the results. The database of certified labels is open-access and can be used by any researcher for scientific purposes.

\section{Preliminaries}
\label{sec:preliminaries}

\subsection{Paroxysmal Atrial Fibrillation}
\label{sec:preliminaries:PxAF}

\begin{figure}[th]
\begin{center}
\centerline{\includegraphics[width=\columnwidth]{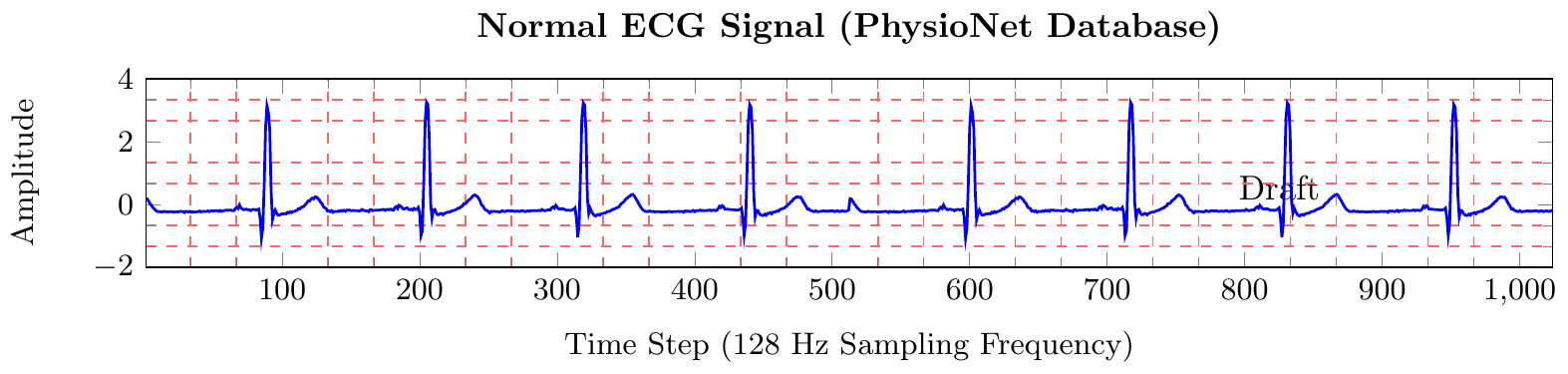}}
\centerline{\includegraphics[width=\columnwidth]{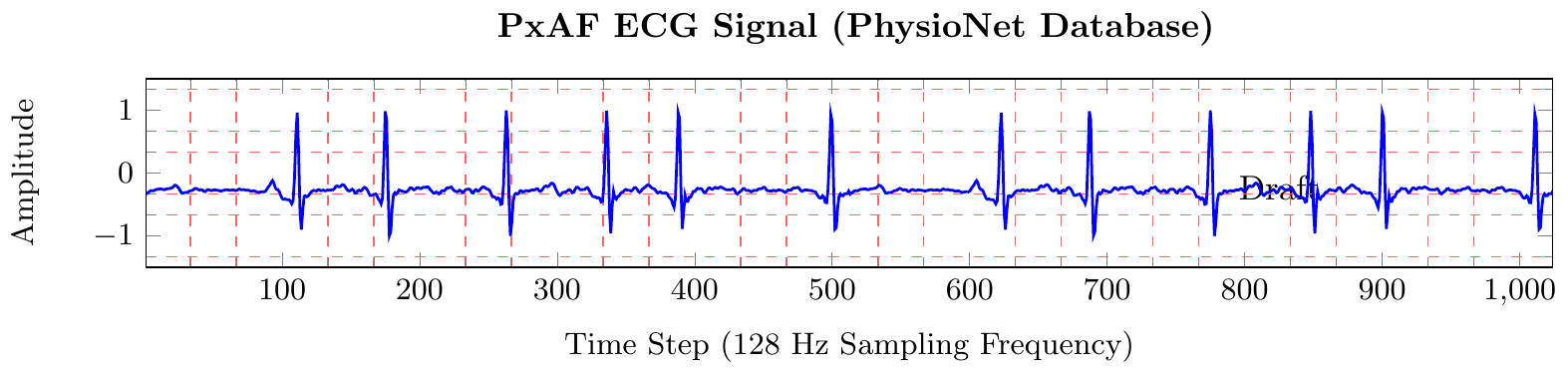}}
\caption {a) Illustration of a sinus rhythm condition. Heart rate variation within 60-100 beats per minute. (b) PxAF condition. Heart rate variability in the form of arrhythmia and P-wave alterations.}
\label{fig:nsr_PxAF}
\end{center}
\end{figure}

ECG is a registration of the electrical activity of heart cells. A normal ECG is a cyclic signal composed of several waves and peaks within each cycle from which the QRS complex, T-wave, and P-wave are mostly regarded as indicative patterns of the signal. Fig.~\ref{fig:nsr_PxAF}.a depicts a normal ECG signal along with the indicative patterns occurring in a certain order in time. The cyclic behavior of the ECG signal comes from the fact that heart muscles have two phases of activity: contraction and relaxation. A contraction is normally followed by a relaxation, where the contraction is initiated from the right atrium down to the ventricles and returned to its initiating point to create a self-stimulating activity through the heart muscles with a rhythmic behavior. This rhythmic action is projected to the ECG signal. The P-wave and QRS complex coincide with the atrial and ventricular contraction, respectively, while the T-wave results from the ventricular relaxation. In the cardiac investigation, a complete relaxation followed by a left ventricle contraction is known as the cardiac cycle. However, for simplicity in ECG signal processing, a cardiac cycle can be defined as the interval between two successive R-peaks for computerized processing.

The morphology of an ECG signal conveys important information about the heart's electrical activity and, to a lesser extent, about its mechanical activity. This includes not only the duration of the QRS complex and the time intervals between the waves and the complex, but also the amplitude of the patterns. Deviation from the typical characteristics of ECG can be resulted either from a physiological condition such as sinus arrhythmia or from pathological conditions, e.g., arrhythmia. Sinus arrhythmia can be dominantly caused by respiration. Paroxysmal Atrial Fibrillation (PxAF) is a pathological condition of the electrical heart action that can happen when the atrial contraction is performed inappropriately. PxAF can initiate an arrhythmia and requires medical considerations and sometimes appropriate management. Fig.~\ref{fig:nsr_PxAF}.b shows a PxAF condition versus a normal sinus rhythm.

%%% figure

As can be seen in Fig.~\ref{fig:nsr_PxAF}, cardiac cycles show a physiological variation of sinus rhythm with clearly visible P-waves in all the cycles. In contrast, in the PxAF case, the P-waves show noticeable alterations over the cycles along with the arrhythmia. An association between PxAF and mortality has been previously demonstrated \cite{Leif2007}. It is also studied that timely detection of PxAF can improve survival in this patient group by appropriate medical management \cite{Leif2007}.

\subsection{Generative Adversarial Networks}
\label{sec:background:GAN}

Generative Adversarial Networks (GANs) are a class of deep learning architectures that have been successfully used to generate synthetic images, time-series data, and other data modalities \cite{jabbar2021survey, cai2021generative}.
In general, GANs are comprised of two sub-networks: the generator ($G$) and the discriminator ($D$). 
$G$ generates synthetic data that is as close as possible to the real data, while $D$ determines whether the generated data is real or not. 
These two sub-networks compete with each other in a two-player minimax game with a loss function of $V(G, D)$ (Eq.~\ref{eq:gan}).
The goal of solving Eq.~\ref{eq:gan} optimization problem is to reach Nash equilibrium \cite{heusel2017gans}.

\begin{equation}
\mathop{min}_{G} \mathop{max}_{D} V (G, D) = E_{x\sim p_{data}(x)} \big[log D(x)\big] + E_{z\sim p_{z}(z)} \big[log (1 - D(G(z)))\big]
\label{eq:gan}
\end{equation}
Probability $D(x)$ determines whether $x$ is generated data or real data.

\section{Related Works}
\label{sec:related_work}

\subsection{PxAF Diagnosis Using DL Methods}
\label{sec:related_work:PxAF_detection}

Previous studies on DL-based methods showed less attention paid to PxAF detection than other forms of arrhythmia \cite{ebrahimi2020review}. Pourbabaee et al. \cite{pourbabaee2018deep} proposed a method for identifying patients with PxAF. Their proposed method employs raw ECG data as input; then, uses a CNN with one fully-connected layer to learn a discriminative pattern of data in the time domain. Plus, they manually tweaked various classification methods to achieve maximum performance. \cite{shashikumar2018detection} proposed an attention-based DL method for detecting PxAF episodes from a synthetic database composed of 24-hour Holter ECG recordings. Time-frequency representations of 30-second windows are fed sequentially into the CNN. Then, the extracted features are presented to a bidirectional recurrent neural network with an attention layer. \cite{gilon2020forecast} constructed a new long-term ECG database (24 to 96 hours) for the purpose of detecting PxAF. After careful analysis by a cardiologist, 250 AF onsets of PxAF have been detected. They proposed a CNN followed by a bidirectional Gated Recurrent Units (GRU) network for PxAF detection. The network was trained to distinguish between RR intervals that precede an AF onset and RR intervals distant from any AF. They concluded that RR intervals contain information about the incoming AF episode. \cite{tzou2021paroxysmal} proposed to predict the occurrence of PxAF by combining wavelet decomposition and a CNN classifier. \cite{surucu2021convolutional} aimed to detect PxAF episodes before occurrence. \cite{surucu2021convolutional} leveraged a CNN to process normalized heart rate variability features resulting in 87.76\% accuracy and 87.50\% f1-score in heart rate variability.

\subsection{Synthetic Data Generation for ECGs}
\label{sec:related_work:data_augmentation}

Medical data tend to be highly sensitive by nature and are often subject to severe usage restrictions. As a result, it is difficult for researchers to collect and share this data. A possible alternative to address the problem of data scarcity is to generate realistic synthetic data \cite{cai2021generative}. \cite{mcsharry2003dynamical, sayadi2010synthetic} proposed mathematical dynamical models to generate continuous ECG signals. These models, however, were limited to one lead signal and did not provide any insight into the mechanism of disease.

Recent studies have demonstrated that GANs are extremely effective at synthesizing ECG waveforms based on a prior distribution of data. Prior works are mainly focused on efficient GAN architecture \cite{delaney2019synthesis, zhu2019electrocardiogram, banerjee2021synthesis, adib2021synthetic, thambawita2021deepfake, li2022tts, xia2023generative}. \cite{delaney2019synthesis} studied various GAN architectures by leveraging  LSTM or BiLSTM as the generator and a CNN discriminator with single or multiple  Convolution-ReLU-Pooling layer(s).  Results show that a BiLSTM GAN with a single \texttt{Convolution-ReLU-Pooling} layer provides the best performance. \cite{zhu2019electrocardiogram} used a BiLSTM-CNN GAN model to generate synthetic ECG signals. A GAN architecture based on a four-layer generator and a five-layer fully-connected discriminator is proposed in \cite{shaker2020generalization}. \cite{banerjee2021synthesis} proposed a multi-GAN method to generate ECG waveforms for atrial fibrillation arrhythmia by combining the output of GAN models. \cite{thambawita2021deepfake} proposed two GAN architectures, WaveGAN$\ast$ and Pulse2Pulse, with the ability to generate synthetic 10-s ECG waveforms. Pulse2Pulse, which is based on a U-net generative model, is superior to producing realistic ECGs. \cite{li2022tts} was the first to propose a transformer-based conditional GAN architecture, named TTS-CGAN, to generate synthetic time-series with sequences of arbitrary length. Compared to popular RNN or LSTM-based GANs for generating time-series \cite{delaney2019synthesis, esteban2017real, yoon2019time}, TTS-CGAN has no difficulties in producing long synthetic sequences. In continuation, \cite{xia2023generative} proposed TCGAN, an architecture combined with a transformer generator and CNN discriminator. 

Despite the success of these methods, they do not guarantee that the generated data is trustworthy, resulting in the failure of classifiers to make accurate predictions. This paper sheds light on the fact that synthesizing high-quality artificial data play a crucial role in accurate predictions. Thus, we propose a novel physician-certified synthetic data generation method that provides ECG samples indistinguishable from real ones. 

\subsection{Neural Architecture Search for ECG}
\label{sec:related_work:nas}

Several DL models have been developed for detecting a variety of cardiac arrhythmias. However, increasing the complexity of manual-designed networks does not always lead to better performance. Moreover, the introduced deep neural networks mostly require a cumbersome phase of trial-and-error, which results in enormous computational costs \cite{loni2020deepmaker}. Recent advances in Neural Architecture Search (NAS) have enabled the designing of scalable and resource-efficient neural architectures. Being inspired by the remarkable success of NAS in the computer vision domain \cite{elsken2019neural}, several techniques very recently proposed to leverage NAS for designing accurate architectures for arrhythmia detection \cite{fayyazifar2020impact, fayyazifar2021accurate, lv2021heart, liu2021automatic, odema2021eexnas}.

Fayyazifar et al. \cite{fayyazifar2020impact} studied the impact of manually tweaking deep neural networks for cardiac abnormality classification. Additionally, they used wavelet decomposition to enhance the classification performance of the PhysioNet Challenge 2020 \cite{alday2020classification}. \cite{fayyazifar2021accurate} employed a  NAS method for AF classification where they achieved an accuracy of 84.15\% on the PhysioNet challenge 2017 \cite{clifford2017af}. Heart-Darts \cite{lv2021heart} proposed a heartbeat classification method by automatically designing an efficient CNN architecture with a differentiable NAS method. Heart-Darts provides state-of-the-art performance, applied to the MIT-BIH arrhythmia database \cite{moody2001impact}. \cite{liu2021automatic} developed a NAS-based learning method to detect cardiovascular diseases in 12-lead ECG data. In particular, they proposed a novel search strategy that optimizes different attention modules of the same network synchronously. EExNAS \cite{odema2021eexnas} designed energy-efficient CNN architectures for detecting Myocardial Infarction (MI) and Human Activity Recognition (HAR) on wearable devices. 

These methods utilize NAS to design an efficient arrhythmia classifier; however, they are limited to optimizing the feature extraction part. Further, it is not conclusive that the findings of the prior studies are reproducible, especially since there is no comprehensive evaluation found in their report \cite{lindauer2020best}.

\section{Methodology}
\label{sec:method}

\subsection{Method Overview}
\label{sec:method:overview}

\begin{figure}[htbp]
\begin{center}
\centerline{\includegraphics[width=\columnwidth]{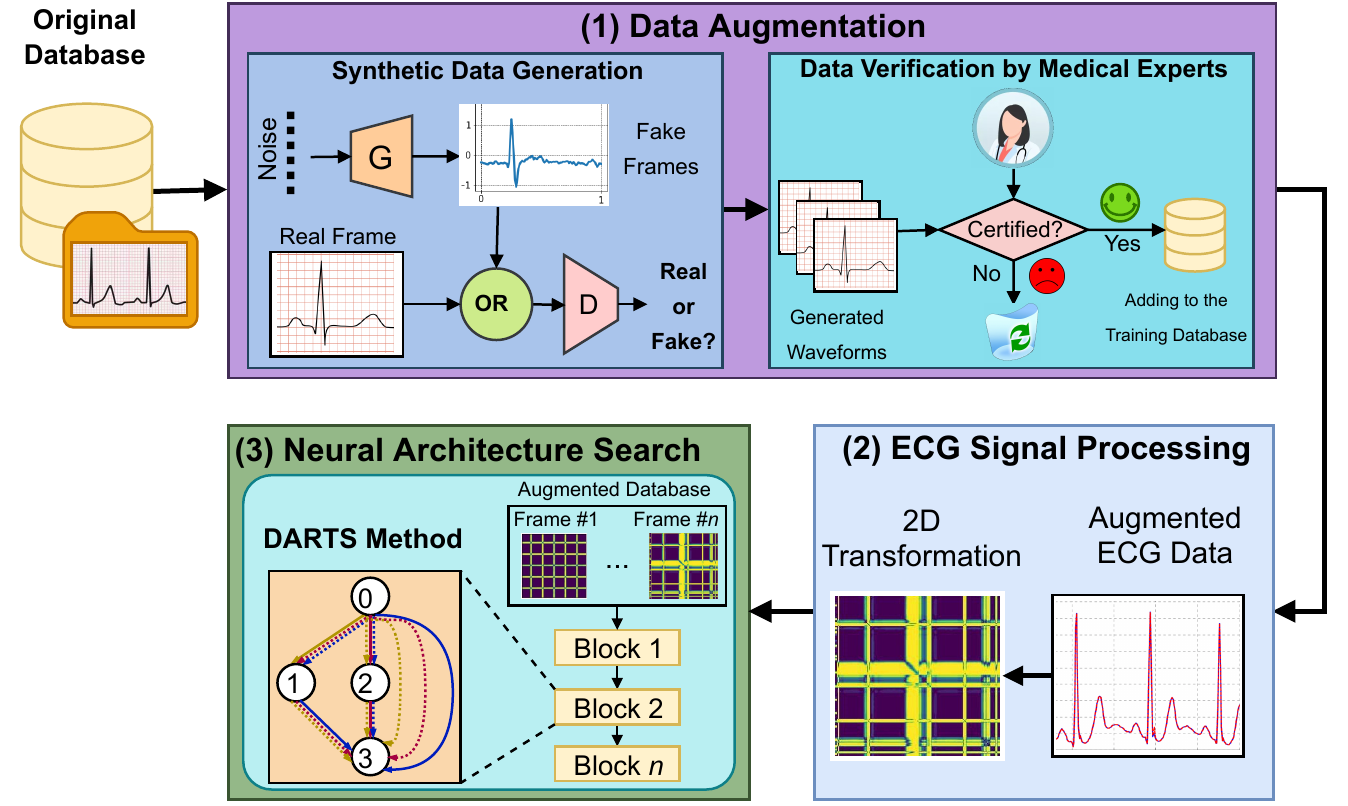}}
\caption{The bird's-eye view of the proposed method.}
\label{fig:method_overview}
\end{center}
\end{figure}

We propose a novel method with three phases, comprising: 1) synthetic data generation, 2) ECG Signal Processing, and 3) CNN Architecture Search. Fig.~\ref{fig:method_overview} depicts the bird's eye view of the proposed method. 
In the first phase, we generate synthetic ECGs for the PxAF class using a GAN model. After the GAN creates synthetic ECGs, an expert physician evaluates them to identify high-quality training data. The second phase of the method employs the wavelet transform of an ECG signal along with the recurrence graph. Rhythmic information of an ECG within short length windows of $4$ second is preserved in a recurrence graph. The outcome of the first stage is a sequence of the two-dimensional images, each incorporating rhythmic contents of a $4$ second interval of an input ECG. In the last phase, a CNN is trained to classify the images where the architecture of the CNN is found using NAS. As we will see, the combination of these innovations noticeably improves the performance of the classification.

\textbf{Phase 1: Certified Synthetic Data Generation.} 
The public databases of ECG mostly contain a heavy class imbalance for the arrhythmia classes. The machine learning methods trained by such databases will be consequently biased for the normal classes. In order to cope with the shortage of signals from the minority class, i.e. the PxAF class, a structure GAN is invoked to create synthetic ECGs from the PxAF class. Obviously, inappropriate synthetic ECGs can mislead the classifier. Therefore, the synthetic ECGs created by the GAN are evaluated by an expert physician in terms of quality using a clearly-defined protocol. The disqualified ECGs will be discarded from the training and the synthetic ECGs certified by the expert physicians will be invoked for the learning process (Section~\ref{sec:method:synthetic_data_generation}).

\textbf{Phase 2: ECG Signal Processing.} ECG signal in its raw form is contaminated by different sources of noises and disturbances, such that the PxAF information can be fully concealed. In order to extract discriminant contents of PxAF from the pathological signals, a level of signal processing is required to purify indicative signal contents (Section~\ref{sec:method:signal_processing}). This processing yields a sequence of 2D images, each containing the dynamics of a few seconds of the signal, to a CNN architecture, in which the ultimate classification is performed.

\textbf{Phase 3: CNN Architecture Search.} Manual design of task-specific neural architectures requires tremendous human effort and domain expertise. In addition, the knowledge learned from designing a network cannot be directly transferred to another person. Neural Architecture Search (NAS) is the process of automatically optimizing a neural network architecture. NAS research has shown significant progress in enabling accurate neural architectures for computer vision applications \cite{loni2021faststereonet, loni2020deepmaker, cai2018proxylessnas, elsken2019neural}. Because of this insight, we came up with the idea of leveraging NAS with the hope of improving the accuracy of PxAF detection (Section~\ref{sec:method:CNN_optimization}).

\subsection{Certified Synthetic Data Generation} 
\label{sec:method:synthetic_data_generation}

\subsubsection{GAN Architecture} 
In this paper, we used the Pulse2Pulse GAN model proposed by \cite{thambawita2021deepfake}. Here, we briefly present generator and discriminator architectures. Then, we present the procedure of certifying the quality of generated data with the help of an expert physician.

\textbf{Generator.} The architecture of the generator is inspired by the U-Net architecture. The U-Net implementation uses 1D convolutional layers for ECG signal generation. The network takes a 2$\times$5000 noise vector to generate a 2-lead signal, which is equal to the dimension of the output layer. The noise is passed through six down-sampling blocks followed by six up-sampling blocks. Each down-sampling block consists of a 1D-convolution layer followed by a Leaky ReLU activation. The deconvolution blocks were built from a series of four layers: an up-sampling layer, a constant padding layer, a 1D-convolution layer, and a ReLU activation function consecutively.  

\textbf{Discriminator.} The discriminator takes an ECG as input and outputs a score indicating how close it is to a fake ECG. The architecture is composed of seven convolutional layers that follow the \texttt{Convolution+Leaky~ReLU+Phase~Shuffle} order. Using phase shuffle operation, each feature map's phase is uniformly perturbed \cite{donahue2018adversarial}. Training specification is reported in Table\ref{tab:configuration}.

\subsubsection{Synthetic Data Certification.}
We observe that not all GAN-generated synthetic ECGs cannot be used as training segments due to their improper morphology, and thus, leveraging all GAN-generated segments in the training process will negatively affect the classification accuracy. Based on the morphological characteristics of ECG signal for PxAF cases, an expert physician manually verified all the synthetic ECGs and certified the valid ones based on the directives listed in  Table~\ref{tab:Expert_Reject}. In this table, the bizare shape implies on the condition in which the sequence of the ECG peaks and wave, and/or their shapes fundamentally differ from the ones, seen in the clinical practice. This condition might be seen in a segment (directive 2), or the entire of the synthetic ECG. It was also observed that the QRS complexes of the synthetic ECG are inconsistent, or accompanied by extra weird morphology (directives 3, 4). The PxAF characteristics were inconsistently seen in some of the data, affecting the learning process, and thus were eliminated (directive 5). 

\begin{table}[htbp]
\centering
\caption{Directives for rejecting improper synthetic ECG segments.}
\label{tab:Expert_Reject}
\resizebox{\textwidth}{!}{%
\begin{tabular}{llc}
\hline
\textbf{Directive} & \textbf{Explanation} & \textbf{Plot}\\ \hline
1. Bizarre Shape       &   Improper morphology with undetectable peaks or waves     &    Fig.~\ref{fig:Expert_Reject}.b   \\
2. Distorted PxAF       &   There are distorted segments of the signal   with bizarre shape  &   Fig.~\ref{fig:Expert_Reject}.c    \\
3. Inconsistent QRS-complex  &  Heart beat exist, but the QRS-complexes are inconsistent in different beats  &   Fig.~\ref{fig:Expert_Reject}.d   \\
4. Redundant/Noisy R  peaks  &  Extra and noisy R peaks in the segment    &    Fig.~\ref{fig:Expert_Reject}.e   \\
5. Partial PxAF       &   Segment partially include PxAF pattern   &   Fig.~\ref{fig:Expert_Reject}.f  \\ \hline
\end{tabular}%
}
\end{table}

\begin{figure}[th]
\begin{center}
\centerline{\includegraphics[width=0.7\columnwidth]{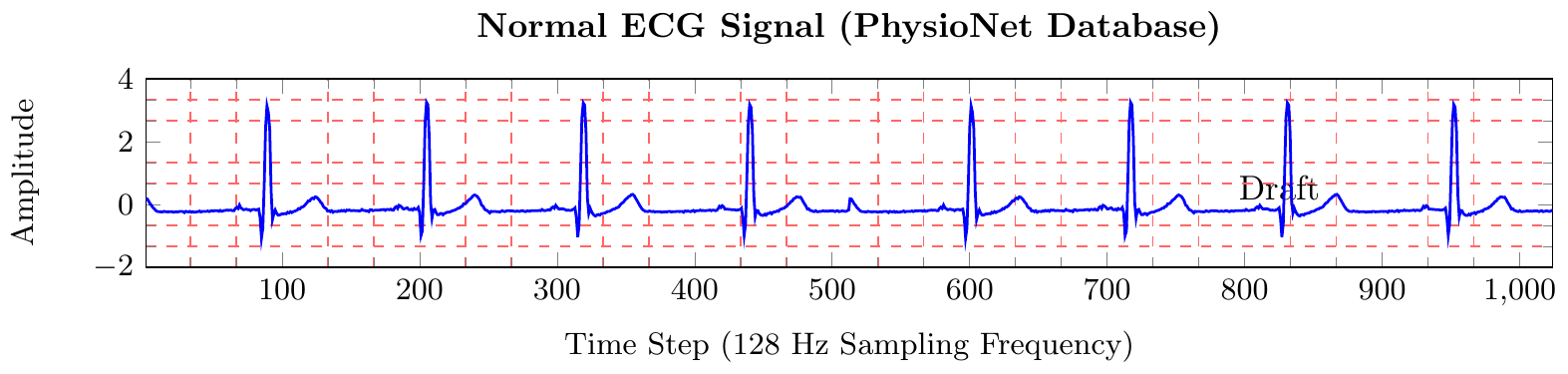}}
\centerline{\includegraphics[width=0.7\columnwidth]{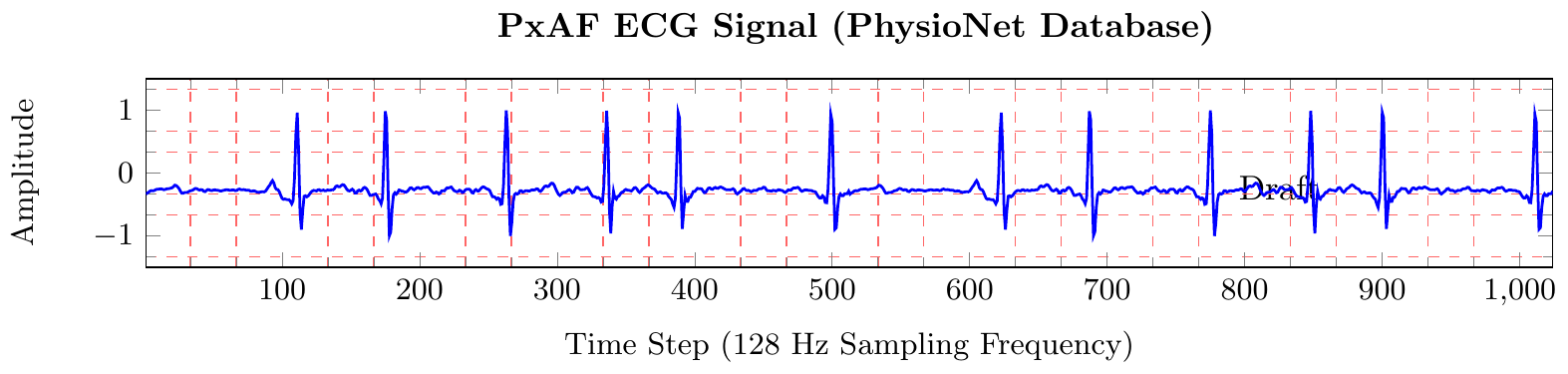}}
\centerline{\includegraphics[width=0.7\columnwidth]{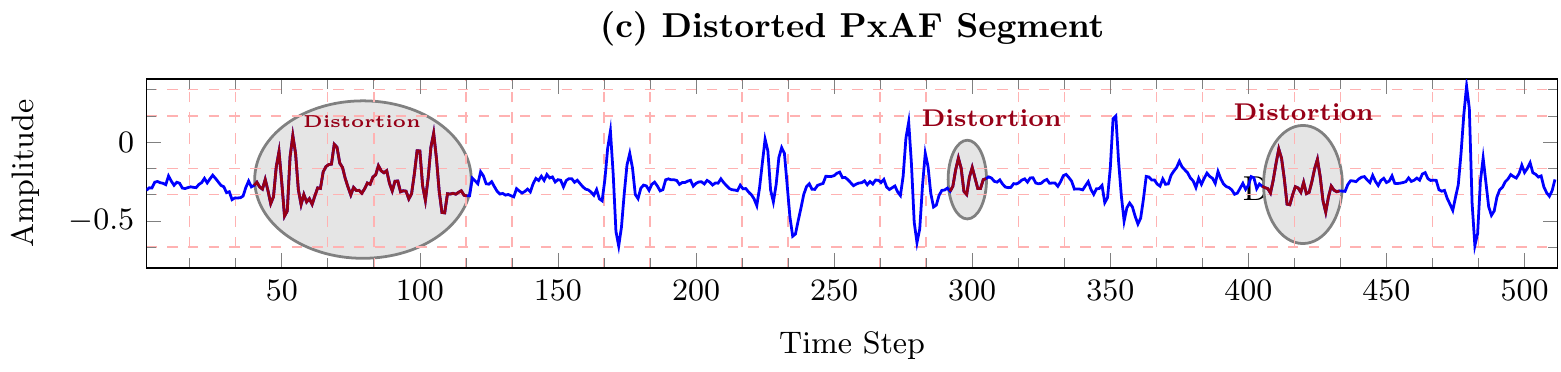}}
\centerline{\includegraphics[width=0.7\columnwidth]{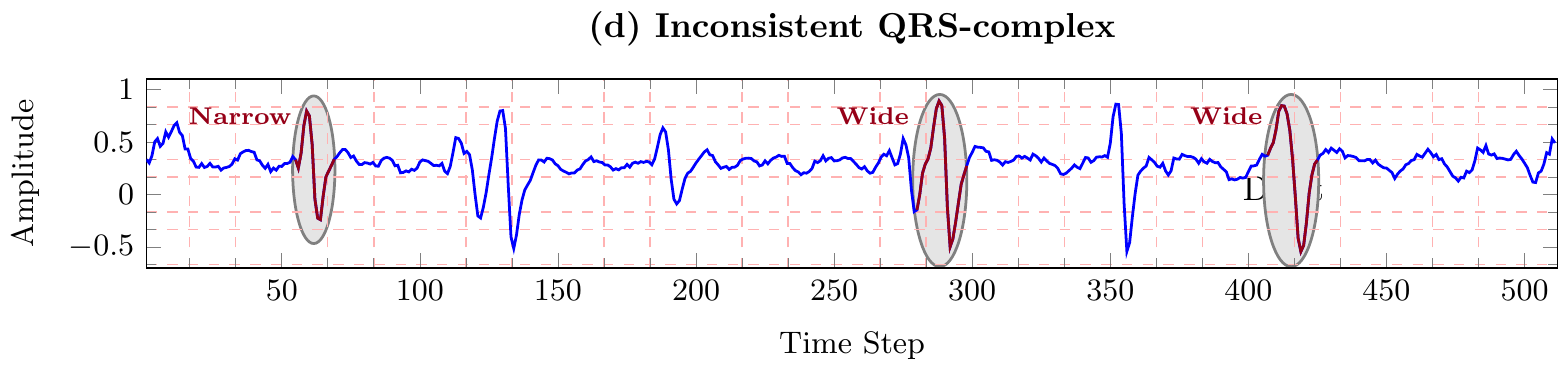}}
\centerline{\includegraphics[width=0.7\columnwidth]{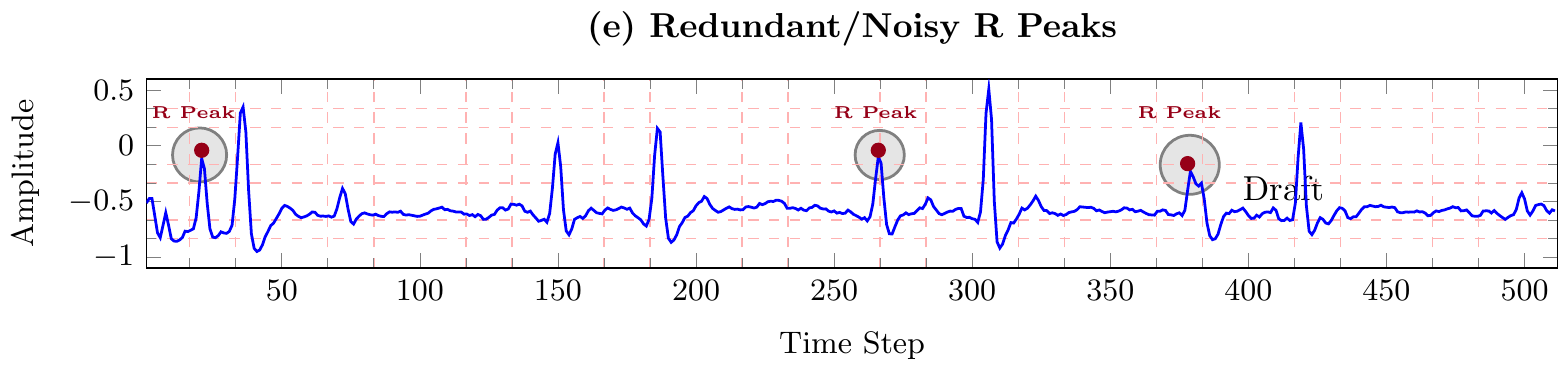}}
\centerline{\includegraphics[width=0.7\columnwidth]{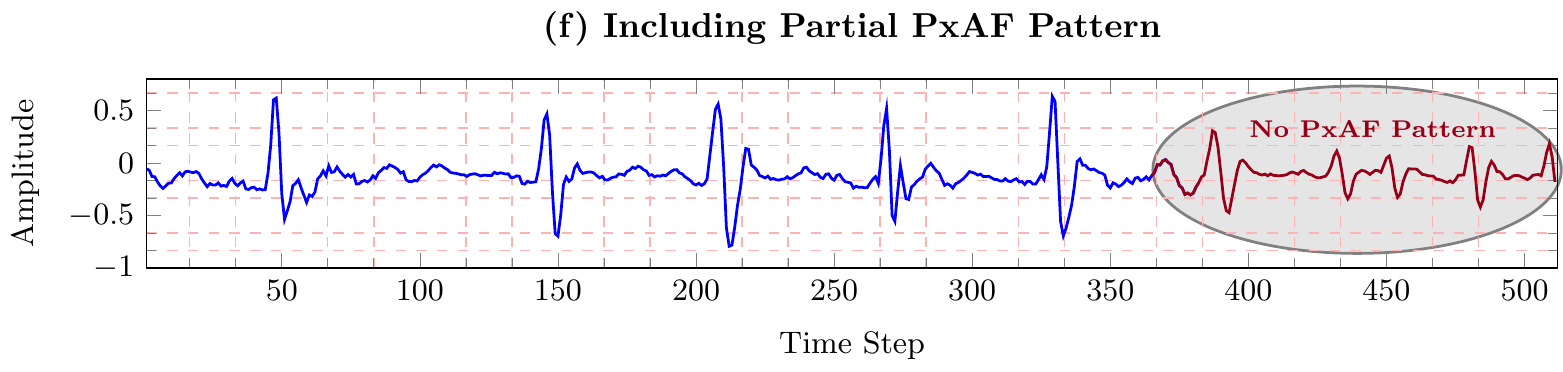}}
\caption {(a) Plotting a certified synthetic PxAF sample. Plotting PxAF synthetic samples rejected by an expert physician due to (b) bizarre shape, (c) distorted PxAF, (d) inconsistent QRS-complex, (e) redundant/noisy R peaks (showing with red points), and (f) partially existing PxAF pattern in the segment. The sampling frequency is 128 Hz.}
\label{fig:Expert_Reject}
\end{center}
\end{figure}

\subsection{ECG Signal Processing}
\label{sec:method:signal_processing}

\begin{figure}[th]
\begin{center}
\centerline{\includegraphics[width=\columnwidth]{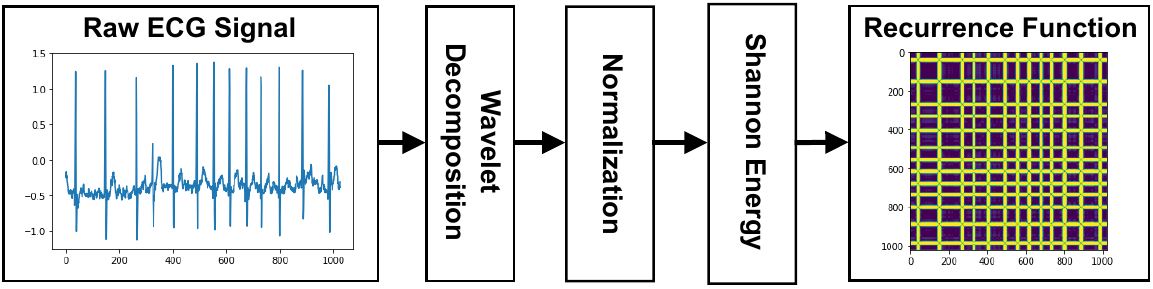}}
\caption{Illustration of the proposed signal processing pipeline.}
\label{fig:signal_processing:1}
\end{center}
\end{figure}

Fig.~\ref{fig:signal_processing:1} shows the major steps of the proposed signal processing pipeline. As shown, the input ECG signal is firstly decomposed to its constitutive components using wavelet transformation until the $10^{th}$ level using the Daubechies 3 wavelet family. The detail of the wavelet transforms at the levels $2,3$ and $4$ along with the approximation contents of the $10*{th}$ level are reconstructed and added together, to eliminate the noises and the disturbances contaminating the signal. The resulting signal is then normalized by the absolute value of the points with the largest value. Next, the Shannon energy of the normalized signal is calculated using the following formula:

\begin{equation}
\textbf{Y}_{i}(t) = x^2(t) log (x^2(t))
\label{eq:1}
\end{equation}

where $x(t)$ is the normalized ECG signal which is positively biased to secure non-zero values. An envelope of the resulting Shannon energy signal is found by using a non-overlapping temporal window of length $100 ms$, which slides over the signal. Lastly, a recurrence 2D function of the envelope is obtained. Calculational details of finding the recurrence plot are found in \cite{vibration2040021}. The output of the signal processing algorithm is a 2D representation of an input signal, which is discriminant for the PxAF and the normal classes. A CNN employs 2D images for classification. 

\subsection{CNN Architecture Search}
\label{sec:method:CNN_optimization}
In general, the learning proficiency of CNNs will be improved by increasing the number of network layers. However, simply stacking the network layers may cause accuracy degradation since the deeper networks will encounter a vanishing/explosion gradient problem. Neural Architecture Search (NAS) methods aim to help engineers to design highly efficient neural networks from scratch \cite{liu2018darts, loni2020deepmaker, loni2021faststereonet}.

The NAS pipeline typically begins with a pre-defined space of network operators. Since the search space is often enormous (e.g., containing $10^{24}$ or even more possible architectures \cite{loni2021faststereonet}), it is unlikely that an exhaustive search is tractable. Thus, heuristic search methods are widely applied to speed up the search process. At an early age, each sampled architecture undergoes an individual training process from scratch, and thus the overall computational overhead is large, e.g., hundreds of GPU-days (e.g., \cite{tan2019mnasnet} requires 3800 GPU days).

To alleviate the computing cost of NAS methods, researchers proposed to share computation among the sampled architectures, with the key idea of reusing network weights trained previously \cite{cai2018efficient, loni2021faststereonet} or starting from a well-trained super-network \cite{pham2018efficient}. These efforts shed light on the one-shot NAS methods, which require training the super-network only once, and therefore run more efficiently (e.g., 2-3 orders of magnitude faster than conventional approaches). 

One-shot NAS methods jointly formulate architecture search and network training \cite{cai2018proxylessnas, liu2018darts, dong2019searching}. Differentiable NAS methods solve this problem using gradient-based algorithms such as Stochastic Gradient Descent (SGD). DARTS \cite{liu2018darts} is a well-known differentiable NAS method that constructs a super-network with all possible operators. DARTS utilizes a cell-based design space to search for a well-behaved cell architecture \cite{dong2019searching, liu2018darts}. Then, the cell may be stacked any number of times to meet various hardware devices' resource requirements. In this paper, we utilize DARTS \cite{liu2018darts} to design CNN architectures due to significantly reducing the notorious design time of neural networks.
 
Mathematically, the final DARTS architecture is a function, $f(x; \omega, \alpha)$, where $x$ is input, $\omega$ is network parameters (e.g., convolutional kernels), and $\alpha$ in architectural parameters (e.g., indicating the importance of each operator between each pair of layers). $f(x; \omega, \alpha)$ is differentiable to both $\omega$ and $\alpha$ could be optimized using the SGD algorithm. $f(x; \omega, \alpha)$ is composed of a few cells, where each cell of DARTS is defined by a directed acyclic graph with a pre-defined number of layers and a limited set of neural operators. Each cell contains $N$ nodes, and there is a predefined set, $E$, which indicates connected pairs of nodes. For each connected node pair $(i, j)$ and $i<j$, node $j$ takes $x_i$ as input and propagates it through a pre-deﬁned operator set, $O$, and sums up all outputs (Eq.~\ref{eq:3}). $O$ supports separable convolution ($3\times3$, $5\times5$), dilated convolution ($3\times3$, $5\times5$), max/average-pooling ($3\times3$), and Identify operators.

\begin{equation}
y^{(i,j)}(x_i)=\sum_{o \in O}\frac{exp(\alpha_{o^{(i,j)}})}{\sum_{o' \in O}exp(\alpha_{o^{(i,j)}}}.o(X_i)
\label{eq:3}
\end{equation}

The normalization is performed by computing the Softmax function over the architectural weights. $\alpha$ and $\omega$ get optimized alternately in each search iteration. Afterward, the operator $o$ with the maximum value is preserved for each edge $(i, j)$, and all other network parameters $\omega$ are discarded. In DARTS, the type of each cell is either a normal cell for feature extraction or a reduction cell for both feature extraction and dimension reduction. After designing the optimal cell, we assemble the final network by stacking 18 normal cells with two reduction cells, where every six normal cells are followed by one reduction cell \cite{Loni2021tas}. Last, the final architecture is re-trained from scratch to fine-tune the network parameters.

\section{Experimental Setup}
\label{sec:setup}

\subsection{Database Preparation}
\label{sec:setup:database}

\ourname identifies individuals who are at risk of PxAF. To this end, we utilized the PhysioNet PxAF prediction challenge database \cite{clifford2017af}. This database includes two-channel ECG recordings. The ECG signals were digitized with a 128 Hz sampling frequency, 16 bits per sample, and nominally 200 A/D units per millivolt. The database is divided into training and testing sets. The original train set consists of 100 records with a duration of 30 minutes that are collected for normal individuals and PAF patients, each with an equal number of recordings. The test set contains 50 records of 30 minutes duration in which 28 subjects are at risk of PxAF, and 22 subjects are healthy individuals. We completely isolate the training and testing sets. We also did not create a separate validation set to evaluate training performance since the size of the database is relatively small.

In this paper, we partitioned each 30 minutes ECG signal into segments of four seconds duration resulting in 512 samples/segment. To build the original database ($D_{Original}$), we randomly select 4231, 906, and 906 segments for train, validation, and testing, respectively. We consider two classes for training and testing sets: normal (healthy) and PxAF patients. $D_{Original}$ contains 3545 and 2498 samples for normal and PxAF classes, respectively. The ECG data labeling tool will be released alongside the codes upon acceptance. 

We generate 10000 synthetic segments for the PxAF class using GAN. As we have data imbalance for the PxAF class, we only synthesize PxAF segments. The original training database has been augmented with 10000 synthetic segments ($D_{GAN}$). Due to the fact that most of the generated segments are not of high quality, an expert physician evaluated all synthetic data and certified 539 segments containing PxAF. Then, we add the certified synthetic PxAF segments to the original training set to make the final synthetic database ($D_{CGAN}$). The synthetic data generation time takes $\approx$ 42 GPU hours on a single NVIDIA\textsuperscript{®} GTX 1080ti that produces $\approx$4.3 Kg $CO_2$ \cite{lacoste2019quantifying}. 

\subsection{Configuration Setup}
\label{sec:setup:configuration_setup}

Table~\ref{tab:configuration} summarizes the configuration setup of experiments. In this paper, each DARTS cell consists of seven nodes equipped with a depth-wise concatenation operation as the output node. The convolutional operations follow the \texttt{Convolution+Batch~Normalization+ReLU} order. The network design time (search+re-training) takes $\approx$ 9 GPU hours on a single NVIDIA\textsuperscript{®} GTX 1080ti that produces $\approx$0.97 Kg $CO_2$ \cite{lacoste2019quantifying}. The rest of the setup follows \cite{liu2018darts}.

\begin{table}[htbp]
\centering
	\caption{The configuration setup of the signal processing and neural architecture search hyper-parameters.}
	\begin{center}
		\resizebox{\columnwidth}{!}{
		\begin{tabular}{cc}
			\hline
	   \textbf{Signal Processing Pipeline} & \textbf{Value}  \\ \hline
			Maximum wavelet scales	& level 10 \\
			Shannon Window Length	& 0.1 second \\ 
			Recurrence Length &  4 second \\ \hline	
		\textbf{Synthetic Data Generation} & \textbf{Value}\\ \hline
	       \# Epochs 	&  8000\\ 
		      Optimizer  &  Adam \\ 
		      Learning Rate ($lr$)  &   1.0$\times$10\textsuperscript{-4}\\ \hline	
		\textbf{NAS Hyper-parameters: Design} & \textbf{Value}\\ \hline
	        Train/Test Segments	& 5000/1000 \\ 
			\# Epochs 	&  50 \\ 
			Batch Size  & 	6  \\
			Optimizer &  SGD \\
			Learning Rate ($lr$)	& 0.025 \\
			weight decay & 3$\times$10\textsuperscript{-4} \\ 
			momentum & 0.9 \\	\hline
		\textbf{NAS Hyper-parameters: Fine-tuning} & \textbf{Value}  \\ \hline
			\# Epochs 	& 200 \\ 
			Batch Size  & 	10  \\
			Optimizer & SGD \\
			Learning Rate ($lr$)	& 0.025 \\
			weight decay & 3.0$\times$10\textsuperscript{-4} \\ 
			momentum & 0.9 \\	\hline	
		\multicolumn{2}{c}{\textbf{Hardware Specification}}    \\ \hline
		    GPU  & NVIDIA\textsuperscript{®} GTX 1080ti (2.5 GHz)\\
            GPU Compiler	&  cuDNN Version 7.1  \\
            Operating System	& Ubuntu 18.04  \\
            Training System Memory & 32 GB \\\hline 
		\end{tabular}}
		\label{tab:configuration}
	\end{center}
\end{table} 

\subsection{Baseline for Comparison}
\label{sec:setup:optimization_baseline}

\textbf{Auto-Sklearn \cite{feurer-arxiv20a}.} Auto-Sklearn is a state-of-the-art library for automated machine learning (AutoML) that is compatible with the scikit-learn library \cite{pedregosa2011scikit}. Auto-Sklearn automatically selects appropriate hyperparameters for a given database by leveraging Bayesian optimization \cite{shahriari2015taking} as the search method. Auto-Sklearn uses four data preprocessing techniques, 14 feature preprocessing techniques, 15 classifiers, and a structured hypothesis space with 110 hyperparameters. Auto-Sklearn considers the past performance of similar databases and constructs ensembles from the machine learning models evaluated during the optimization to improve the optimization quality. Due to the high efficiency of Auto-Sklearn in customizing the machine learning pipeline \cite{conrad2022benchmarking, ribeiro2022benchmarking}, we consider Auto-Sklearn as the second comparison baseline.

\textbf{Deep Residual Network (ResNet) \cite{he2016deep}.} ResNet is a family of handcrafted architectures that won the ILSVRC competition challenge in 2015. ResNet is constructed by several back-to-back residual blocks connected to a final linear fully-connected layer. In this study, we used ResNet as the third comparison baseline since ResNet has been widely used in automated clinical diagnosis of various diseases \cite{shi2020review, bhattacharjya2021existing, yeh2022deep, ebrahimi2020review}.

\subsection{Performance Measurement}
\label{sec:setup:measurement}

This section introduces common quantitative metrics used for presenting how well synthetic data generation and classification methods work.

\textbf{GAN Performance.} For evaluating the performance of GAN, we use a database containing GAN output data and original data to train a model, which is then tested on a held-out set of true examples. This requires the generated data to have labels - an expert physician provides labels to GAN output data. We statistically analyze the distribution of read ECGs and fake ECGs using Kolmogorov-Smirnov test (K-S test). Plus, we will show the Q-Q plot to look at the skewness of fake data from real data.

\textbf{Classifier Performance.} The formulas for quantifying measurements are listed below:

\begin{equation}
Accuracy=\frac{TP + TN}{TP + TN + FP + FN}
\end{equation}

\begin{equation}
Specificity=\frac{TN}{TN + FP}
\end{equation}
\begin{equation}
Sensitivity=\frac{TP}{TP + FN}
\end{equation}

where $TP$, $TN$, $FP$, and $FN$ denote True Positives, True Negatives, False Positive, and False Negative, respectively.

\section{Experimental Results}
\label{sec:results}

\subsection{The Synthetic ECGs}
\label{sec:results:Data_Augmentation}
\begin{figure}[th]
\begin{center}
\centerline{\includegraphics[width=\columnwidth]{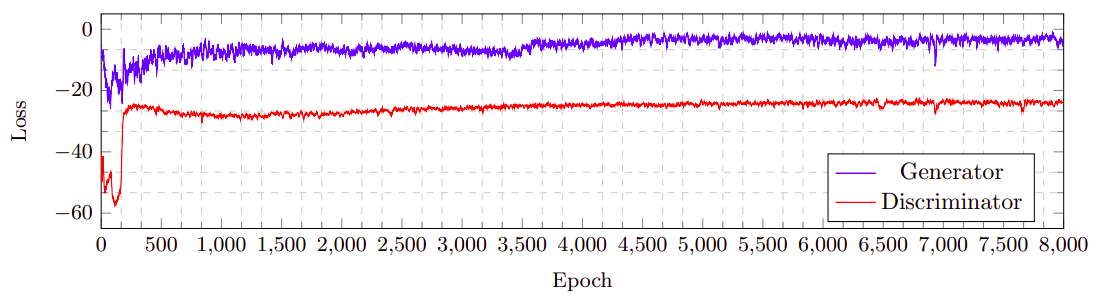}}
\caption {Loss of the discriminator and generator during GAN training.}
\label{fig:GAN_Trainin}
\end{center}
\end{figure}

The previously-described GAN is trained with 8000 epochs and a learning rate of $0.0001$. Fig.~\ref{fig:GAN_Trainin} depicts the loss function of the generator and the discriminator of the GAN. Both of the losses converge to a similar low margin implying the learning relevance. The outcomes of the GAN generator constitute our synthetic ECGs.

The quality of the synthetic ECGs is evaluated based on the statistical measures, separately applied to the entire original and synthetic populations, once using the outcomes of the certified-GAN and once using the GAN without accreditation of the expert physician. In both cases, the fidelity of the synthetic ECGs is evaluated by using the two PxAF-related parameters of ECG: heart rate and R-peak to R-peak interval (RR Interval). These two parameters are independently calculated for the populations using the signal processing algorithm described in Section \ref{sec:method:signal_processing}. It is worth noting that these two parameters reflect the variation of the cardiac cycle and heart rate that is linked to arrhythmia. 

\begin{figure}[th]
\begin{center}
\centerline{\includegraphics[width=\columnwidth]{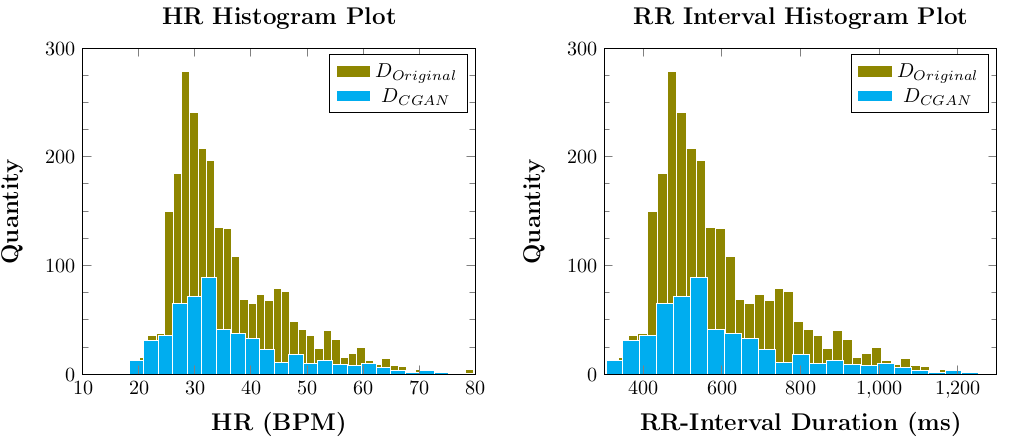}}
\caption {Distribution of (left) heart rates and (right) RR interval in all 539 certified segments ($D_{CGAN}$) compared to the original database ($D_{Original}$).}
\label{fig:Histogram}
\end{center}
\end{figure}

In total, 10000 synthetic ECGs were generated using the previously-described GAN, from which 539 were accredited by the expert physician. Fig.~\ref{fig:Histogram} illustrates the histogram of the two PxAF-related parameters for the real and the synthetic ECGs resulting from the certified-GAN. The modal similarities are obviously seen for the synthetic and real populations.

In order to explore the fidelity of the synthetic ECGs, descriptive statistics are calculated over the three populations: The real ECGs, the GAN, and the certified-GAN subjected to having PxAF condition. Table~\ref{tab:Data_Augmentation} represents the mean, standard deviation, and percentile values corresponding to the three populations. From the population perspective, the two PxAF-related parameters of the certified synthetic ECGs demonstrate very good fitness to the population of the real ECGs, with a marginal deviation of less than 2\% for the mean value. This value is almost 4\% for the data from the GAN. The deviation of the percentile values is less than 10\%. The certified-GAN provides clear improvements in all the statistics, but the 2.5\% percentile which corresponds to the outlier data.

\begin{table}[htbp]
\centering
\caption{Mean, standard deviation (STD), 2.5\%, and 97.5\% percentile for HR and RR interval parameters in real and synthetic ECGs. BPM stands for beats per minute. }
\label{tab:Data_Augmentation}
\resizebox{\textwidth}{!}{%
\begin{tabular}{c|cccc|cccc|cccc}
\hline
 \multirow{3}{*}{\textbf{Feature}}&  \multicolumn{12}{c}{\textbf{Database}}   \\ 
 &
  \multicolumn{4}{c}{\textbf{$D_{Original}$}} &
  \multicolumn{4}{c}{\textbf{$D_{GAN}$}} &
  \multicolumn{4}{c}{\textbf{$D_{CGAN}$}} \\ \cline{2-13}
 &
  \textbf{Mean} & \textbf{STD} & \textbf{2.5\%} & \textbf{97.5\%} & \textbf{Mean} & \textbf{STD} & \textbf{2.5\%} & \textbf{97.5\%} & \textbf{Mean} & \textbf{STD} & \textbf{2.5\%} & \textbf{97.5\%} \\ \hline
     
\textbf{RR Interval (ms)} & 597.6 & 164.0 & 390.6 & 1007.8 & 621.01 & 218.75 & 376.0  & 1164.06 & 604.29 & 203.12 & 351.56 & 1101.56 \\

\textbf{HR (BPM)} & 35.86 & 10  & 23.43  & 60.46 & 37.25 & 13 & 21.56 & 69.84 & 36.26  & 12 & 21.09 & 66.09 \\ \hline
\end{tabular}%
}
\end{table}

\begin{figure}[H]
\centering
\captionsetup{justification=centering}
\resizebox{\columnwidth}{!}{
\begin{tabular}{cc}
\centerline{\includegraphics[width=\columnwidth]{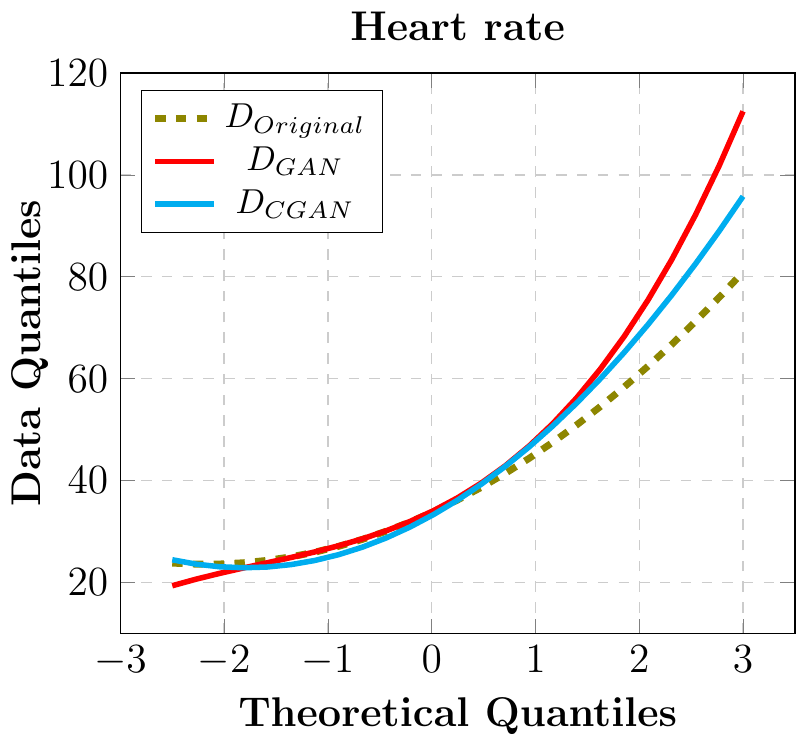}}
&
\centerline{\includegraphics[width=\columnwidth]{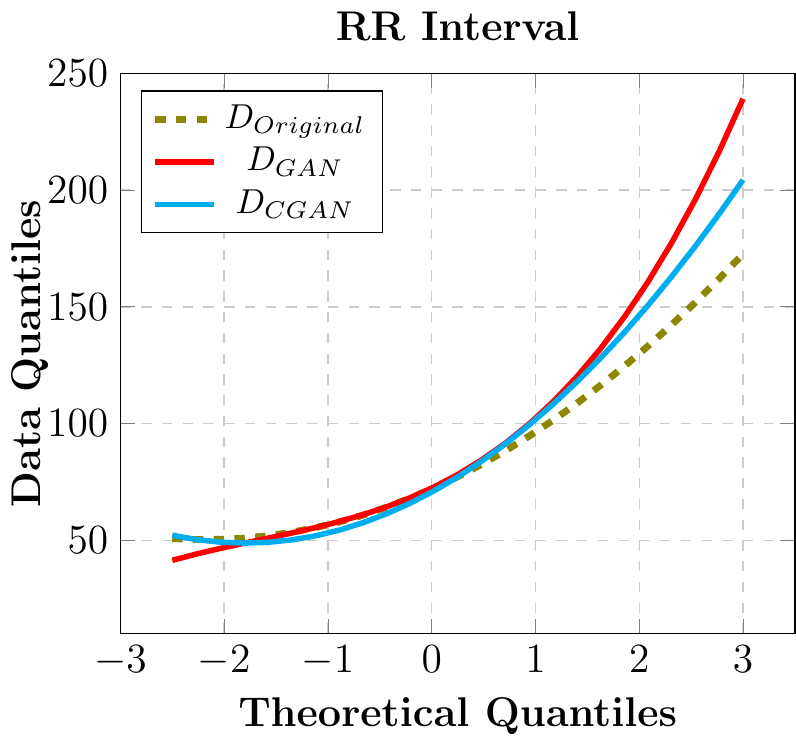}}
\end{tabular}
}
\caption {Illustration of the \textit{Q-Q}-plot for (left) heart rate, and (right) RR interval. }
\label{fig:Q-Q_Plot}
\end{figure}

In order to obtain a better understanding of the outperformance of the certified-GAN, the quantile distribution of the real and synthetic data, the so-called Q-Q plot, is investigated. Fig.~\ref{fig:Q-Q_Plot} illustrates the resulting Q-Q plot. 

It is obviously seen that the certified-GAN provides closer statistical distribution to the real one, as compared to the plain GAN. This is also explored by using the Kolmogorov-Smirnov Test.

Table~\ref{tab:K-S_test:RR-HR} presents the results of the Kolmogorov-Smirnov (K-S) test for heart rate. As seen in the table, the certified-GAN improves the K-S statistics as well as the p-value, showing a closer distribution to the real population. This distribution is closer to the real population than the one for the GAN, confirming the effectiveness of the certified-GAN.

\begin{table}[H]
\centering
\caption{Kolmogorov-Smirnov Test Results.}
\label{tab:K-S_test:RR-HR}
\resizebox{0.9\textwidth}{!}{%
\begin{tabular}{c|ccc}
\hline
  \multirow{2}{*}{\textbf{Parameter}}&  \multicolumn{3}{c}{\textbf{Database}}   \\ 
 & (\textbf{$D_{Original}$} \& \textbf{$D_{GAN}$}) & (\textbf{$D_{Original}$} \& \textbf{$D_{CGAN}$})& (\textbf{$D_{CGAN}$} \& \textbf{$D_{GAN}$}) \\ \hline
Statistic  & 0.0593 & 0.0758 & 0.04631   \\
p-value &  1.4897e-06 &  0.0116 & 0.2157\\                      \hline
\end{tabular}%
}
\end{table}

\subsection{PxAF Classification Performance}
\label{sec:results:PxAF_Classification}

Table~\ref{tab:PxAF_Classification} compares the results of \ourname with the state-of-the-art and state-of-practice classification methods. Results show that \ourname provides the most accurate classification result with 99\% accuracy compared to all counterparts. 

\begin{table}[ht]
\caption{Comparing the results of \ourname  with state-of-the-art and state-of-practice methods.}
\label{tab:PxAF_Classification}
\resizebox{\columnwidth}{!}{
\begin{tabular}{ccc}
\hline
\textbf{Method}  &  \multicolumn{2}{c}{\textbf{PhysioNet Classification Accuracy (\%)} }   \\  \hline

Pourbabaee et al. \cite{pourbabaee2018deep}$\ddagger$ & \multicolumn{2}{c}{91.0} \\

Surucu et al. \cite{surucu2021convolutional}&   \multicolumn{2}{c}{93.88} \\ \hline 

 &   \textbf{ $D_{Original}$ (\%)} & \textbf{ $D_{CGAN}$ (\%)}   \\ \hline 

ResNet-18 \cite{he2016deep} &   95.2 & 97.0\\

Auto\_Sklearn \cite{feurer-arxiv20a} &   92.53 & 92.83\\

\ourname (Ours)  &   \textbf{\textcolor{blue}{ 97.3}} & \textbf{\textcolor{blue}{ 99.0}}\\ \hline

\multicolumn{3}{c}{$\dagger$ Using the same search space as DARTS. }	\\
\multicolumn{3}{c}{$\ddagger$ Reporting the best results by CNN architecture with a K-nearest neighbor (KNN) classifier. } \\
\hline
\end{tabular}}
\end{table} 

This study proposed an accurate method for screening PxAF. In this application, the trade-off between sensitivity and specificity is made by assigning the threshold of the output layer, where sensitivity and specificity are defined as:

\begin{itemize}
    \item Sensitivity is the probability of PxAF condition when the classification result is positive 
    \item Specificity is the probability of normal condition when the classification result is negative 
\end{itemize}

\begin{figure}[th]
\begin{center}
\centerline{\includegraphics[width=0.5\columnwidth]{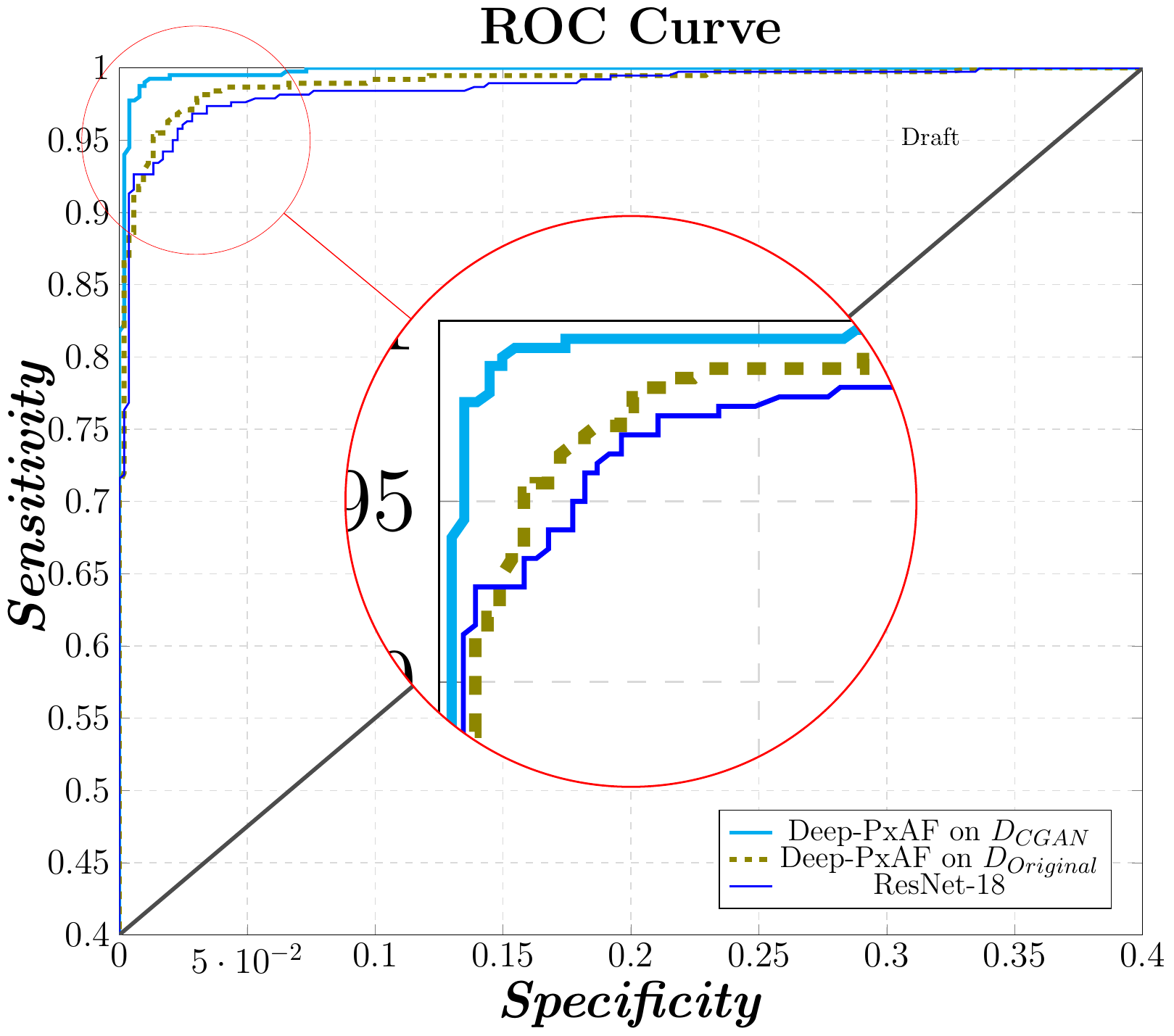}}
\caption {Comparing the ROC curve of \ourname trained on $D_{Original}$ and $D_{CGAN}$ to the ResNet-18 trained on $D_{CGAN}$ baseline method. }
\label{fig:ROC}
\end{center}
\end{figure}

Receiver Operating Characteristics (ROC) is a plot of the $Sensitivity$ against $(1 \mhyphen Specificity)$, in which the optimal point is the point with maximal Sensitivity and specificity. Fig.~\ref{fig:ROC} illustrates the ROC curve for the proposed method in comparison with the ResNet-18 classification method. As can be seen in Fig.~\ref{fig:ROC}, \ourname provides a better characteristic in terms of the compromise between Sensitivity and Specificity with a closer curve to the ideal case of the straight angle. The Area Under the Curve (AUC) of ROC for \ourname trained on $D_{CGAN}$ is improved by 0.32\% and 0.47\% compared to \ourname trained on $D_{Original}$ and ResNet-18 trained on $D_{CGAN}$, respectively.

\subsection{Qualitative Analysis of the Searched Cells}
\label{sec:results:cell_architecture_analysis} 

Fig.~\ref{fig:cell_architecture_analysis} depicts the best cells searched by \ourname for the $D_{CGAN}$ dataset. For the normal cell, DARTS tends to increase the portion of dilated convolution separable convolution (\texttt{sep\_conv}) operations with the 5$\times$5 kernel size. This is because larger kernel sizes (5$\times$5) improve the representational power of the network. In contrast, the reduction cell has many average pooling operations for compressing the information across the spatial dimension. This is because pooling operations can increase the nonlinear representation ability of the network. Referring to the recurrence graphs in which rhythmic contents of ECG are preserved within the squares of $4$ second (see Fig\ref{fig:method_overview}, one can intuitively understand that an optimal kernel size is one that can include rhythms. A small kernel size can negatively impact the learning quality due to its failure to incorporate rhythmic content.

\begin{table}[H]
\centering
\captionsetup{justification=centering}
\resizebox{\columnwidth}{!}{
\begin{tabular}{c}
\centerline{\includegraphics[width=\columnwidth]{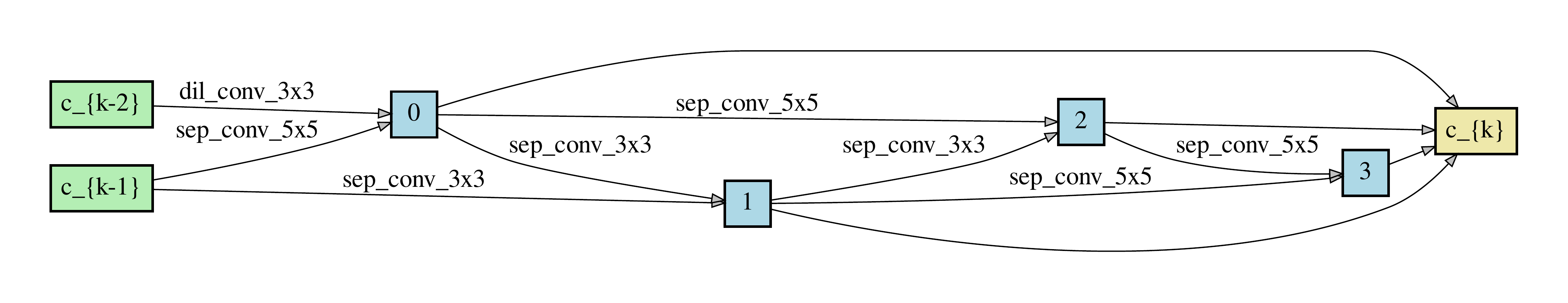}}\\
\textbf{ \large (a)}
\\
\centerline{\includegraphics[width=0.3\columnwidth]{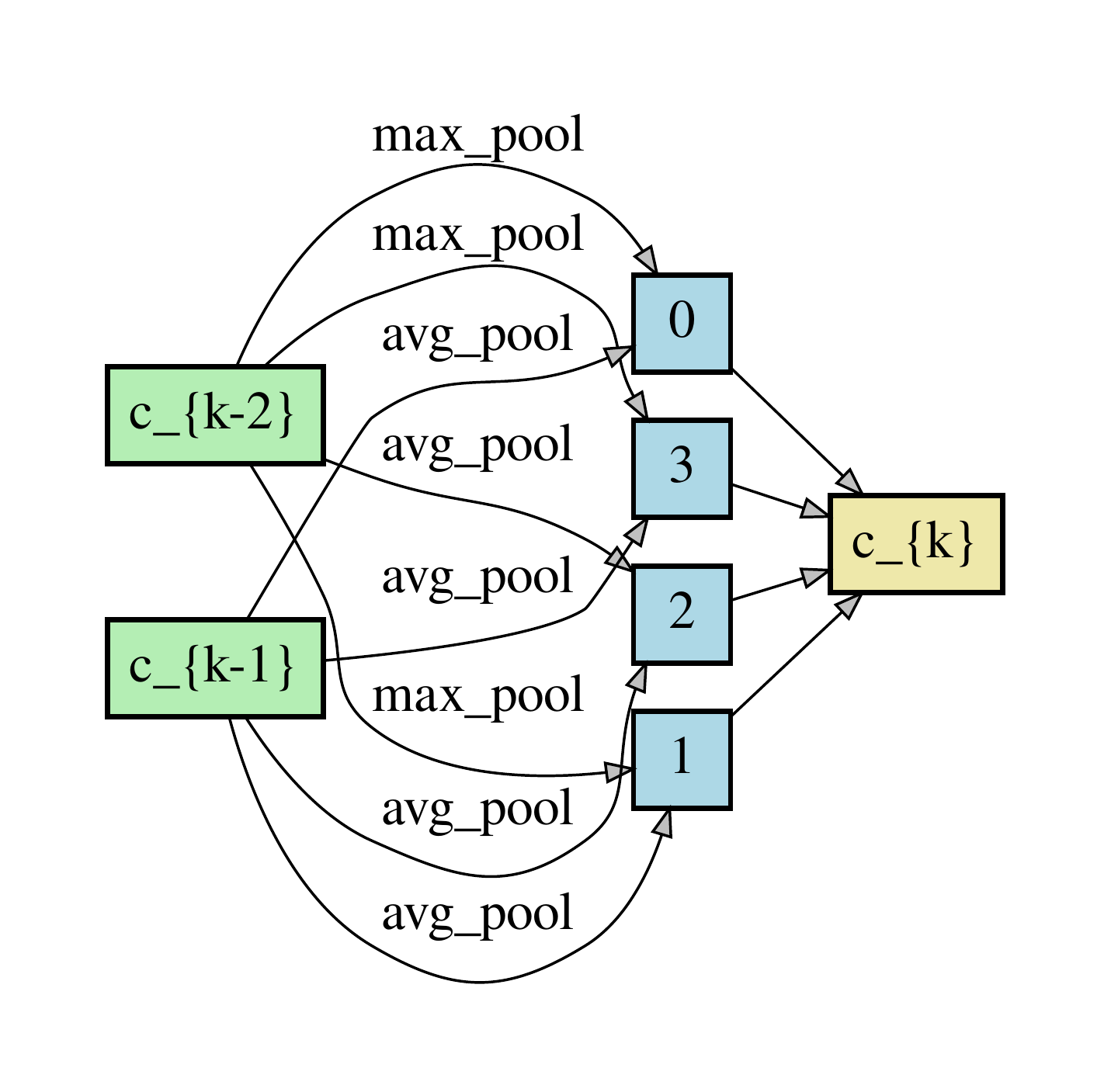}}
\\
\textbf{\large (b)}
\end{tabular}}
\captionof{figure}{(a) Normal cell. (b) Reduction cell.} 
\label{fig:cell_architecture_analysis}
\end{table}

\section{Discussion}
\label{sec:discussion}

This study suggested an original framework for PxAF classification using a novel combination of a GAN and NAS in conjunction with an advanced signal processing method. The plays an important role in enriching the training data by generating valid synthetic ECGs through a certified procedure, and the NAS acts as a reliable architecture designer to boost the classification performance. The resulting classification method was optimized and implemented to detect patients with PxAF arrhythmia, which is regarded as an important case study with vital importance. The proposed method improved the screening accuracy by 6.1\% compared to the state-of-the-art automated machine learning method \cite{feurer-arxiv20a}. The baseline for comparison was ResNet-18 and Auto-Sklearn which are well-known benchmarks for the machine learning method. These benchmarks was noticeably outperformed by the proposed method.

\textbf{Synthetic Data Generation.}
This study employed a GAN architecture to generate synthetic ECGs and meanwhile invoked an expert physician to accredit the synthetic data. The application of GAN in generating synthetic ECG has been already explored \cite{thambawita2021deepfake, zhu2019electrocardiogram}, however, the effectiveness of the generated ECGs in the training process is questionable since inappropriate synthetic data can evidently mislead the classifier. The certified-GAN which was  proposed by this study effectively pruned the inappropriate signals. Results showed a noticeable improvement in the learning process using the certified-GAN. We will make these synthetic signals publicly available to any researcher to explore for any scientific purposes. 

Another interesting aspect of this study is the statistical techniques employed to study the fidelity of synthetic ECGs. Heart rate and R-R interval were employed as the measures for the PxAF. The statistical techniques mainly perform population-based evaluations which fit well into the scope of the learning process. The certified-GAN showed incapability to generate appropriate outliers, as reflected by the 2.5 percentile in Table\ref{tab:Data_Augmentation}. Such outlier data cannot play an important role in the learning process performed by the proposed deep learning architecture.

\textbf{ECG Signal Processing.} 
In this study, the rhythmic contents of the heartbeats are innovatively preserved at the feature extraction level through signal processing and the recurrence images. Like other methods sufficing to the temporal features, there are a number of design parameters associated with the method at this level, such as the window's length for obtaining the recurrence graph as well as the wavelet transformation. These parameters were empirically obtained based on prior knowledge of the signal. Integration of finding the optimal values for these design parameters with the optimization process might provide further improvements. 

\textbf{CNN Architecture Search.}
Although several NAS methods have been proposed to detect various arrhythmias \cite{fayyazifar2020impact, fayyazifar2021accurate, lv2021heart, liu2021automatic, odema2021eexnas}, the area is still unexplored for designing an efficient method for PxAF detection based on an optimized architecture of CNN. Moreover, the optimization process was not performed at the feature extraction level. 

We learned the dynamic variation of the heartbeats at the feature learning level by designing customized architectures for recurrence images. Several design parameters are associated with the method at this level, such as the number of training epochs. We empirically obtained these parameters based on prior knowledge about the neural architecture search. PxAF yields higher performance compared to the results of conventional machine learning techniques that are automatically tuned by Auto-Sklearn. This primarily results from our custom-designed CNN architecture's higher feature extraction performance. On the other hand, manually tuning a generic CNN architecture \cite{pourbabaee2018deep} may result in lower accuracy in comparison with Auto-Sklearn.

\textbf{Statement of Reproducibility.} To foster reproducibility:

\begin{itemize}
    \item \textbf{Reproducibility analysis.} Many works on NAS have issues regarding reproducibility due to intrinsic stochasticity. Our project, codes, and labeled datasets will be open-sourced upon acceptance to ensure reproducibility.
    
    \item \textbf{Availability of database.} In this study, we evaluated our networks using the PhysioNet PxAF database \cite{clifford2017af}. Thus, this work does not involve any new data collection or human subject evaluation. The generated ECGs with the corresponding ground truth labels can be downloaded after paper acceptance.
\end{itemize}

\section{Conclusion \& Future Work}
\label{sec:conclusion}

This paper suggested an original combination of certified synthetic data generation in conjunction with the NAS method for classifying a vital pathological sign of ECG signal: Paroxysmal Atrial Fibrillation (PxAF). To overcome privacy and ethical concerns for data sharing, a GAN model was used to generate synthetic data. The synthetic ECGs were purified by an expert physician to discard the irrelevant ones. We employed a CNN for the classification, for which the optimal was found by the NAS. The input images to the CNN were extracted from the ECGs using recurrence graphs of the wavelet transform. It is found that the proposed framework offers a noticeable improvement in classification performance compared to the state-of-the-art as well as the existing benchmarks. In future work, the performance of the classifier resulting from this study will be practically explored on the general population after being implemented in an appropriate platform of wearable ECG. 

%___________________________________________________________________________________________________________________________________________________
	%\clearpage

% \bibliography{references}

\begin{thebibliography}{60}
\providecommand{\natexlab}[1]{#1}
\providecommand{\url}[1]{\texttt{#1}}
\expandafter\ifx\csname urlstyle\endcsname\relax
  \providecommand{\doi}[1]{doi: #1}\else
  \providecommand{\doi}{doi: \begingroup \urlstyle{rm}\Url}\fi

\bibitem[Adib et~al.(2021)Adib, Afghah, and Prevost]{adib2021synthetic}
Edmond Adib, Fatemeh Afghah, and John~J Prevost.
\newblock Synthetic ecg signal generation using generative neural networks.
\newblock \emph{arXiv preprint arXiv:2112.03268}, 2021.

\bibitem[Alday et~al.(2020)Alday, Gu, Shah, Robichaux, Wong, Liu, Liu, Rad,
  Elola, Seyedi, et~al.]{alday2020classification}
Erick A~Perez Alday, Annie Gu, Amit~J Shah, Chad Robichaux, An-Kwok~Ian Wong,
  Chengyu Liu, Feifei Liu, Ali~Bahrami Rad, Andoni Elola, Salman Seyedi, et~al.
\newblock Classification of 12-lead ecgs: the physionet/computing in cardiology
  challenge 2020.
\newblock \emph{Physiological measurement}, 41\penalty0 (12):\penalty0 124003,
  2020.

\bibitem[Banerjee and Ghose(2021)]{banerjee2021synthesis}
Rohan Banerjee and Avik Ghose.
\newblock Synthesis of realistic ecg waveforms using a composite generative
  adversarial network for classification of atrial fibrillation.
\newblock In \emph{2021 29th European Signal Processing Conference (EUSIPCO)},
  pages 1145--1149. IEEE, 2021.

\bibitem[Bhattacharjya and Sarma(2021)]{bhattacharjya2021existing}
Upasana Bhattacharjya and Kandarpa~Kumar Sarma.
\newblock Existing methods and emerging trends for novel coronavirus (covid-19)
  detection using residual network (resnet): A review on deep learning
  analysis.
\newblock \emph{Smart Healthcare Monitoring Using IoT with 5G}, pages 131--147,
  2021.

\bibitem[Cai et~al.(2018{\natexlab{a}})Cai, Chen, Zhang, Yu, and
  Wang]{cai2018efficient}
Han Cai, Tianyao Chen, Weinan Zhang, Yong Yu, and Jun Wang.
\newblock Efficient architecture search by network transformation.
\newblock In \emph{Proceedings of the AAAI Conference on Artificial
  Intelligence}, volume~32, 2018{\natexlab{a}}.

\bibitem[Cai et~al.(2018{\natexlab{b}})Cai, Zhu, and Han]{cai2018proxylessnas}
Han Cai, Ligeng Zhu, and Song Han.
\newblock Proxylessnas: Direct neural architecture search on target task and
  hardware.
\newblock \emph{arXiv preprint arXiv:1812.00332}, 2018{\natexlab{b}}.

\bibitem[Cai et~al.(2021)Cai, Xiong, Xu, Wang, Li, and Pan]{cai2021generative}
Zhipeng Cai, Zuobin Xiong, Honghui Xu, Peng Wang, Wei Li, and Yi~Pan.
\newblock Generative adversarial networks: A survey toward private and secure
  applications.
\newblock \emph{ACM Computing Surveys (CSUR)}, 54\penalty0 (6):\penalty0 1--38,
  2021.

\bibitem[Clifford et~al.(2017)Clifford, Liu, Moody, Li-wei, Silva, Li, Johnson,
  and Mark]{clifford2017af}
Gari~D Clifford, Chengyu Liu, Benjamin Moody, H~Lehman Li-wei, Ikaro Silva,
  Qiao Li, AE~Johnson, and Roger~G Mark.
\newblock Af classification from a short single lead ecg recording: The
  physionet/computing in cardiology challenge 2017.
\newblock In \emph{2017 Computing in Cardiology (CinC)}, pages 1--4. IEEE,
  2017.

\bibitem[Conrad et~al.(2022)Conrad, M{\"a}lzer, Schwarzenberger, Wiemer, and
  Ihlenfeldt]{conrad2022benchmarking}
Felix Conrad, Mauritz M{\"a}lzer, Michael Schwarzenberger, Hajo Wiemer, and
  Steffen Ihlenfeldt.
\newblock Benchmarking automl for regression tasks on small tabular data in
  materials design.
\newblock \emph{Scientific Reports}, 12\penalty0 (1):\penalty0 1--14, 2022.

\bibitem[Delaney et~al.(2019)Delaney, Brophy, and Ward]{delaney2019synthesis}
Anne~Marie Delaney, Eoin Brophy, and Tomas~E Ward.
\newblock Synthesis of realistic ecg using generative adversarial networks.
\newblock \emph{arXiv preprint arXiv:1909.09150}, 2019.

\bibitem[Donahue et~al.(2018)Donahue, McAuley, and
  Puckette]{donahue2018adversarial}
Chris Donahue, Julian McAuley, and Miller Puckette.
\newblock Adversarial audio synthesis.
\newblock \emph{arXiv preprint arXiv:1802.04208}, 2018.

\bibitem[Dong and Yang(2019)]{dong2019searching}
Xuanyi Dong and Yi~Yang.
\newblock Searching for a robust neural architecture in four gpu hours.
\newblock In \emph{Proceedings of the IEEE/CVF Conference on Computer Vision
  and Pattern Recognition}, pages 1761--1770, 2019.

\bibitem[Ebrahimi et~al.(2020)Ebrahimi, Loni, Daneshtalab, and
  Gharehbaghi]{ebrahimi2020review}
Zahra Ebrahimi, Mohammad Loni, Masoud Daneshtalab, and Arash Gharehbaghi.
\newblock A review on deep learning methods for ecg arrhythmia classification.
\newblock \emph{Expert Systems with Applications: X}, 7:\penalty0 100033, 2020.

\bibitem[Elsken et~al.(2019)Elsken, Metzen, and Hutter]{elsken2019neural}
Thomas Elsken, Jan~Hendrik Metzen, and Frank Hutter.
\newblock Neural architecture search: A survey.
\newblock \emph{The Journal of Machine Learning Research}, 20\penalty0
  (1):\penalty0 1997--2017, 2019.

\bibitem[Esteban et~al.(2017)Esteban, Hyland, and R{\"a}tsch]{esteban2017real}
Crist{\'o}bal Esteban, Stephanie~L Hyland, and Gunnar R{\"a}tsch.
\newblock Real-valued (medical) time series generation with recurrent
  conditional gans.
\newblock \emph{arXiv preprint arXiv:1706.02633}, 2017.

\bibitem[Fayyazifar(2021)]{fayyazifar2021accurate}
Najmeh Fayyazifar.
\newblock An accurate cnn architecture for atrial fibrillation detection using
  neural architecture search.
\newblock In \emph{2020 28th European Signal Processing Conference (EUSIPCO)},
  pages 1135--1139. IEEE, 2021.

\bibitem[Fayyazifar et~al.(2020)Fayyazifar, Ahderom, Suter, Maiorana, and
  Dwivedi]{fayyazifar2020impact}
Najmeh Fayyazifar, Selam Ahderom, David Suter, Andrew Maiorana, and Girish
  Dwivedi.
\newblock Impact of neural architecture design on cardiac abnormality
  classification using 12-lead ecg signals.
\newblock In \emph{2020 Computing in Cardiology}, pages 1--4. IEEE, 2020.

\bibitem[Feurer et~al.(2020)Feurer, Eggensperger, Falkner, Lindauer, and
  Hutter]{feurer-arxiv20a}
Matthias Feurer, Katharina Eggensperger, Stefan Falkner, Marius Lindauer, and
  Frank Hutter.
\newblock Auto-sklearn 2.0: Hands-free automl via meta-learning.
\newblock \emph{arXiv:2007.04074 [cs.LG]}, 2020.

\bibitem[Friberg et~al.(2007)Friberg, Hammar, Pettersson, and
  Rosenqvist]{Leif2007}
Leif Friberg, Niklas Hammar, Hans Pettersson, and Ma\aa~rten Rosenqvist.
\newblock Increased mortality in paroxysmal atrial fibrillation: report from
  the stockholm cohort-study of atrial fibrillation (scaf).
\newblock \emph{Eur. Heart J.}, 28(19):\penalty0 2346--53, 2007.

\bibitem[Gharehbaghi and Lindén(2018)]{Gharehbaghi-ITNNLS}
Arash Gharehbaghi and Maria Lindén.
\newblock A deep machine learning method for classifying cyclic time series of
  biological signals using time-growing neural network.
\newblock \emph{IEEE Transactions on Neural Networks and Learning Systems},
  29\penalty0 (9):\penalty0 4102--4115, 2018.
\newblock \href{https://doi.org/10.1109/TNNLS.2017.2754294}{\ttfamily\path{
  doi:10.1109/TNNLS.2017.2754294}}.

\bibitem[Gharehbaghi et~al.(2017)Gharehbaghi, Sepehri, Lindén, and
  Babic]{Gharehbaghi2018-VSD}
Arash Gharehbaghi, Amir~A Sepehri, Maria Lindén, and Ankica Babic.
\newblock Intelligent phonocardiography for screening ventricular septal defect
  using time growing neural network.
\newblock In \emph{Studies in health technology and informatics}, volume 238,
  pages 108--111. IOC Press, 2017.

\bibitem[Gharehbaghi et~al.(2019)Gharehbaghi, Lindén, and
  Babic]{Gharehbaghi-ASC}
Arash Gharehbaghi, Maria Lindén, and Ankica Babic.
\newblock An artificial intelligent-based model for detecting systolic
  pathological patterns of phonocardiogram based on time-growing neural
  network.
\newblock \emph{Applied Soft Computing}, 83:\penalty0 105615, 2019.
\newblock ISSN 1568-4946.
\newblock
  \href{https://doi.org/https://doi.org/10.1016/j.asoc.2019.105615}{\ttfamily\path{
  doi:https://doi.org/10.1016/j.asoc.2019.105615}}.
\newblock URL
  \url{https://www.sciencedirect.com/science/article/pii/S1568494619303953}.

\bibitem[Gilon et~al.(2020)Gilon, Gr{\'e}goire, and Bersini]{gilon2020forecast}
C{\'e}dric Gilon, Jean-Marie Gr{\'e}goire, and Hugues Bersini.
\newblock Forecast of paroxysmal atrial fibrillation using a deep neural
  network.
\newblock In \emph{2020 International Joint Conference on Neural Networks
  (IJCNN)}, pages 1--7. IEEE, 2020.

\bibitem[Goncalves et~al.(2020)Goncalves, Ray, Soper, Stevens, Coyle, and
  Sales]{goncalves2020generation}
Andre Goncalves, Priyadip Ray, Braden Soper, Jennifer Stevens, Linda Coyle, and
  Ana~Paula Sales.
\newblock Generation and evaluation of synthetic patient data.
\newblock \emph{BMC medical research methodology}, 20\penalty0 (1):\penalty0
  1--40, 2020.

\bibitem[Goswami(2019)]{vibration2040021}
Bedartha Goswami.
\newblock A brief introduction to nonlinear time series analysis and recurrence
  plots.
\newblock \emph{Vibration}, 2\penalty0 (4):\penalty0 332--368, 2019.
\newblock ISSN 2571-631X.
\newblock \href{https://doi.org/10.3390/vibration2040021}{\ttfamily\path{
  doi:10.3390/vibration2040021}}.
\newblock URL \url{https://www.mdpi.com/2571-631X/2/4/21}.

\bibitem[He et~al.(2016)He, Zhang, Ren, and Sun]{he2016deep}
Kaiming He, Xiangyu Zhang, Shaoqing Ren, and Jian Sun.
\newblock Deep residual learning for image recognition.
\newblock In \emph{Proceedings of the IEEE conference on computer vision and
  pattern recognition}, pages 770--778, 2016.

\bibitem[Heusel et~al.(2017)Heusel, Ramsauer, Unterthiner, Nessler, and
  Hochreiter]{heusel2017gans}
Martin Heusel, Hubert Ramsauer, Thomas Unterthiner, Bernhard Nessler, and Sepp
  Hochreiter.
\newblock Gans trained by a two time-scale update rule converge to a local nash
  equilibrium.
\newblock \emph{Advances in neural information processing systems}, 30, 2017.

\bibitem[Jabbar et~al.(2021)Jabbar, Li, and Omar]{jabbar2021survey}
Abdul Jabbar, Xi~Li, and Bourahla Omar.
\newblock A survey on generative adversarial networks: Variants, applications,
  and training.
\newblock \emph{ACM Computing Surveys (CSUR)}, 54\penalty0 (8):\penalty0 1--49,
  2021.

\bibitem[Khushi et~al.(2021)Khushi, Shaukat, Alam, Hameed, Uddin, Luo, Yang,
  and Reyes]{khushi2021comparative}
Matloob Khushi, Kamran Shaukat, Talha~Mahboob Alam, Ibrahim~A Hameed, Shahadat
  Uddin, Suhuai Luo, Xiaoyan Yang, and Maranatha~Consuelo Reyes.
\newblock A comparative performance analysis of data resampling methods on
  imbalance medical data.
\newblock \emph{IEEE Access}, 9:\penalty0 109960--109975, 2021.

\bibitem[Lacoste et~al.(2019)Lacoste, Luccioni, Schmidt, and
  Dandres]{lacoste2019quantifying}
Alexandre Lacoste, Alexandra Luccioni, Victor Schmidt, and Thomas Dandres.
\newblock Quantifying the carbon emissions of machine learning.
\newblock \emph{arXiv preprint arXiv:1910.09700}, 2019.

\bibitem[Li et~al.(2022)Li, Ngu, and Metsis]{li2022tts}
Xiaomin Li, Anne Hee~Hiong Ngu, and Vangelis Metsis.
\newblock Tts-cgan: A transformer time-series conditional gan for biosignal
  data augmentation.
\newblock \emph{arXiv preprint arXiv:2206.13676}, 2022.

\bibitem[Lindauer and Hutter(2020)]{lindauer2020best}
Marius Lindauer and Frank Hutter.
\newblock Best practices for scientific research on neural architecture search.
\newblock \emph{Journal of Machine Learning Research}, 21\penalty0
  (243):\penalty0 1--18, 2020.

\bibitem[Liu et~al.(2018)Liu, Simonyan, and Yang]{liu2018darts}
Hanxiao Liu, Karen Simonyan, and Yiming Yang.
\newblock Darts: Differentiable architecture search.
\newblock \emph{arXiv preprint arXiv:1806.09055}, 2018.

\bibitem[Liu et~al.(2021)Liu, Wang, Gao, and Shi]{liu2021automatic}
Zuhao Liu, Huan Wang, Yibo Gao, and Shunchen Shi.
\newblock Automatic attention learning using neural architecture search for
  detection of cardiac abnormality in 12-lead ecg.
\newblock \emph{IEEE Transactions on Instrumentation and Measurement},
  70:\penalty0 1--12, 2021.

\bibitem[Loni et~al.(2020)Loni, Sinaei, Zoljodi, Daneshtalab, and
  Sj{\"o}din]{loni2020deepmaker}
Mohammad Loni, Sima Sinaei, Ali Zoljodi, Masoud Daneshtalab, and Mikael
  Sj{\"o}din.
\newblock Deepmaker: A multi-objective optimization framework for deep neural
  networks in embedded systems.
\newblock \emph{Microprocessors and Microsystems}, 73:\penalty0 102989, 2020.

\bibitem[Loni et~al.(2021)Loni, Zoljodi, Majd, Ahn, Daneshtalab, Sj{\"o}din,
  and Esmaeilzadeh]{loni2021faststereonet}
Mohammad Loni, Ali Zoljodi, Amin Majd, Byung~Hoon Ahn, Masoud Daneshtalab,
  Mikael Sj{\"o}din, and Hadi Esmaeilzadeh.
\newblock Faststereonet: A fast neural architecture search for improving the
  inference of disparity estimation on resource-limited platforms.
\newblock \emph{IEEE Transactions on Systems, Man, and Cybernetics: Systems},
  2021.

\bibitem[Loni et~al.(2022)Loni, Mousavi, Riazati, Daneshtalab, and
  Sj{\"o}din]{Loni2021tas}
Mohammad Loni, Hamid Mousavi, Mohammad Riazati, Masoud Daneshtalab, and Mikael
  Sj{\"o}din.
\newblock Tas:ternarized neural architecture search for resource-constrained
  edge devices.
\newblock In \emph{Design, Automation \& Test in Europe Conference \&
  Exhibition DATE'22, 14 March 2022, Antwerp, Belgium}. IEEE, March 2022.
\newblock URL \url{http://www.es.mdh.se/publications/6351-}.

\bibitem[Lv et~al.(2021)Lv, Ye, Sun, Zhao, and Lv]{lv2021heart}
Jindi Lv, Qing Ye, Yanan Sun, Juan Zhao, and Jiancheng Lv.
\newblock Heart-darts: Classification of heartbeats using differentiable
  architecture search.
\newblock \emph{arXiv preprint arXiv:2105.00693}, 2021.

\bibitem[McSharry et~al.(2003)McSharry, Clifford, Tarassenko, and
  Smith]{mcsharry2003dynamical}
Patrick~E McSharry, Gari~D Clifford, Lionel Tarassenko, and Leonard~A Smith.
\newblock A dynamical model for generating synthetic electrocardiogram signals.
\newblock \emph{IEEE transactions on biomedical engineering}, 50\penalty0
  (3):\penalty0 289--294, 2003.

\bibitem[Moody and Mark(2001)]{moody2001impact}
George~B Moody and Roger~G Mark.
\newblock The impact of the mit-bih arrhythmia database.
\newblock \emph{IEEE Engineering in Medicine and Biology Magazine}, 20\penalty0
  (3):\penalty0 45--50, 2001.

\bibitem[Odema et~al.(2021)Odema, Rashid, and Al~Faruque]{odema2021eexnas}
Mohanad Odema, Nafiul Rashid, and Mohammad~Abdullah Al~Faruque.
\newblock Eexnas: Early-exit neural architecture search solutions for low-power
  wearable devices.
\newblock In \emph{2021 IEEE/ACM International Symposium on Low Power
  Electronics and Design (ISLPED)}, pages 1--6. IEEE, 2021.

\bibitem[Ogawa et~al.(2018)Ogawa, An, Ikeda, Aono, Doi, Ishii, Iguchi,
  Masunaga, Esato, Tsuji, Wada, Hasegawa, Abe, Lip, Akao, and null
  null]{Hisashi2018}
Hisashi Ogawa, Yoshimori An, Syuhei Ikeda, Yuya Aono, Kosuke Doi, Mitsuru
  Ishii, Moritake Iguchi, Nobutoyo Masunaga, Masahiro Esato, Hikari Tsuji,
  Hiromichi Wada, Koji Hasegawa, Mitsuru Abe, Gregory~Y.H. Lip, Masaharu Akao,
  and null null.
\newblock Progression from paroxysmal to sustained atrial fibrillation is
  associated with increased adverse events.
\newblock \emph{Stroke}, 49\penalty0 (10):\penalty0 2301--2308, 2018.
\newblock \href{https://doi.org/10.1161/STROKEAHA.118.021396}{\ttfamily\path{
  doi:10.1161/STROKEAHA.118.021396}}.
\newblock
  \href{http://arxiv.org/abs/https://www.ahajournals.org/doi/pdf/10.1161/STROKEAHA.118.021396}{{\ttfamily
  https://www.ahajournals.org/doi/pdf/10.1161/STROKEAHA.118.021396}}.
\newblock URL
  \url{https://www.ahajournals.org/doi/abs/10.1161/STROKEAHA.118.021396}.

\bibitem[Pedregosa et~al.(2011)Pedregosa, Varoquaux, Gramfort, Michel, Thirion,
  Grisel, Blondel, Prettenhofer, Weiss, Dubourg, et~al.]{pedregosa2011scikit}
Fabian Pedregosa, Ga{\"e}l Varoquaux, Alexandre Gramfort, Vincent Michel,
  Bertrand Thirion, Olivier Grisel, Mathieu Blondel, Peter Prettenhofer, Ron
  Weiss, Vincent Dubourg, et~al.
\newblock Scikit-learn: Machine learning in python.
\newblock \emph{the Journal of machine Learning research}, 12:\penalty0
  2825--2830, 2011.

\bibitem[Pham et~al.(2018)Pham, Guan, Zoph, Le, and Dean]{pham2018efficient}
Hieu Pham, Melody Guan, Barret Zoph, Quoc Le, and Jeff Dean.
\newblock Efficient neural architecture search via parameters sharing.
\newblock In \emph{International Conference on Machine Learning}, pages
  4095--4104. PMLR, 2018.

\bibitem[Pourbabaee et~al.(2018)Pourbabaee, Roshtkhari, and
  Khorasani]{pourbabaee2018deep}
Bahareh Pourbabaee, Mehrsan~Javan Roshtkhari, and Khashayar Khorasani.
\newblock Deep convolutional neural networks and learning ecg features for
  screening paroxysmal atrial fibrillation patients.
\newblock \emph{IEEE Transactions on Systems, Man, and Cybernetics: Systems},
  48\penalty0 (12):\penalty0 2095--2104, 2018.

\bibitem[Ribeiro et~al.(2022)Ribeiro, Orzechowski, Wagenaar, and
  Moore]{ribeiro2022benchmarking}
Pedro~Henrique Ribeiro, Patryk Orzechowski, Joost Wagenaar, and Jason~H Moore.
\newblock Benchmarking automl algorithms on a collection of binary problems.
\newblock \emph{arXiv preprint arXiv:2212.02704}, 2022.

\bibitem[Sayadi et~al.(2010)Sayadi, Shamsollahi, and
  Clifford]{sayadi2010synthetic}
Omid Sayadi, Mohammad~B Shamsollahi, and Gari~D Clifford.
\newblock Synthetic ecg generation and bayesian filtering using a gaussian
  wave-based dynamical model.
\newblock \emph{Physiological measurement}, 31\penalty0 (10):\penalty0 1309,
  2010.

\bibitem[Shahriari et~al.(2015)Shahriari, Swersky, Wang, Adams, and
  De~Freitas]{shahriari2015taking}
Bobak Shahriari, Kevin Swersky, Ziyu Wang, Ryan~P Adams, and Nando De~Freitas.
\newblock Taking the human out of the loop: A review of bayesian optimization.
\newblock \emph{Proceedings of the IEEE}, 104\penalty0 (1):\penalty0 148--175,
  2015.

\bibitem[Shaker et~al.(2020)Shaker, Tantawi, Shedeed, and
  Tolba]{shaker2020generalization}
Abdelrahman~M Shaker, Manal Tantawi, Howida~A Shedeed, and Mohamed~F Tolba.
\newblock Generalization of convolutional neural networks for ecg
  classification using generative adversarial networks.
\newblock \emph{IEEE Access}, 8:\penalty0 35592--35605, 2020.

\bibitem[Shashikumar et~al.(2018)Shashikumar, Shah, Clifford, and
  Nemati]{shashikumar2018detection}
Supreeth~P Shashikumar, Amit~J Shah, Gari~D Clifford, and Shamim Nemati.
\newblock Detection of paroxysmal atrial fibrillation using attention-based
  bidirectional recurrent neural networks.
\newblock In \emph{Proceedings of the 24th ACM SIGKDD International Conference
  on Knowledge Discovery \& Data Mining}, pages 715--723, 2018.

\bibitem[Shi et~al.(2020)Shi, Wang, Shi, Wu, Wang, Tang, He, Shi, and
  Shen]{shi2020review}
Feng Shi, Jun Wang, Jun Shi, Ziyan Wu, Qian Wang, Zhenyu Tang, Kelei He,
  Yinghuan Shi, and Dinggang Shen.
\newblock Review of artificial intelligence techniques in imaging data
  acquisition, segmentation, and diagnosis for covid-19.
\newblock \emph{IEEE reviews in biomedical engineering}, 14:\penalty0 4--15,
  2020.

\bibitem[Surucu et~al.(2021)Surucu, Isler, Perc, and
  Kara]{surucu2021convolutional}
M~Surucu, Y~Isler, M~Perc, and R~Kara.
\newblock Convolutional neural networks predict the onset of paroxysmal atrial
  fibrillation: Theory and applications.
\newblock \emph{Chaos: An Interdisciplinary Journal of Nonlinear Science},
  31\penalty0 (11):\penalty0 113119, 2021.

\bibitem[Tan et~al.(2019)Tan, Chen, Pang, Vasudevan, Sandler, Howard, and
  Le]{tan2019mnasnet}
Mingxing Tan, Bo~Chen, Ruoming Pang, Vijay Vasudevan, Mark Sandler, Andrew
  Howard, and Quoc~V Le.
\newblock Mnasnet: Platform-aware neural architecture search for mobile.
\newblock In \emph{Proceedings of the IEEE/CVF Conference on Computer Vision
  and Pattern Recognition}, pages 2820--2828, 2019.

\bibitem[Thambawita et~al.(2021)Thambawita, Isaksen, Hicks, Ghouse, Ahlberg,
  Linneberg, Grarup, Ellervik, Olesen, Hansen, et~al.]{thambawita2021deepfake}
Vajira Thambawita, Jonas~L Isaksen, Steven~A Hicks, Jonas Ghouse, Gustav
  Ahlberg, Allan Linneberg, Niels Grarup, Christina Ellervik, Morten~Salling
  Olesen, Torben Hansen, et~al.
\newblock Deepfake electrocardiograms using generative adversarial networks are
  the beginning of the end for privacy issues in medicine.
\newblock \emph{Scientific reports}, 11\penalty0 (1):\penalty0 1--8, 2021.

\bibitem[Tucker et~al.(2020)Tucker, Wang, Rotalinti, and Myles]{NPJ2020}
Allan Tucker, Zhenchen Wang, Ylenia Rotalinti, and Puja Myles.
\newblock Generating high-fidelity synthetic patient data for assessing machine
  learning healthcare software.
\newblock \emph{npj Digital Medicine}, 3\penalty0 (1):\penalty0 147, 2020.

\bibitem[Tzou et~al.(2021)Tzou, Lin, and Chen]{tzou2021paroxysmal}
Heng-An Tzou, Shien-Fong Lin, and Peng-Sheng Chen.
\newblock Paroxysmal atrial fibrillation prediction based on morphological
  variant p-wave analysis with wideband ecg and deep learning.
\newblock \emph{Computer Methods and Programs in Biomedicine}, 211:\penalty0
  106396, 2021.

\bibitem[Xia et~al.(2023)Xia, Xu, Chen, Zhang, and Zhang]{xia2023generative}
Yi~Xia, Yangyang Xu, Peng Chen, Jun Zhang, and Yongliang Zhang.
\newblock Generative adversarial network with transformer generator for
  boosting ecg classification.
\newblock \emph{Biomedical Signal Processing and Control}, 80:\penalty0 104276,
  2023.

\bibitem[Yeh et~al.(2022)Yeh, Zhang, Chen, Liu, Wang, Yang, Yeh, Cheng, Chen,
  and Su]{yeh2022deep}
Lee-Ren Yeh, Yang Zhang, Jeon-Hor Chen, Yan-Lin Liu, An-Chi Wang, Jie-Yu Yang,
  Wei-Cheng Yeh, Chiu-Shih Cheng, Li-Kuang Chen, and Min-Ying Su.
\newblock A deep learning-based method for the diagnosis of vertebral fractures
  on spine mri: retrospective training and validation of resnet.
\newblock \emph{European Spine Journal}, pages 1--9, 2022.

\bibitem[Yoon et~al.(2019)Yoon, Jarrett, and Van~der Schaar]{yoon2019time}
Jinsung Yoon, Daniel Jarrett, and Mihaela Van~der Schaar.
\newblock Time-series generative adversarial networks.
\newblock \emph{Advances in neural information processing systems}, 32, 2019.

\bibitem[Zhu et~al.(2019)Zhu, Ye, Fu, Liu, and Shen]{zhu2019electrocardiogram}
Fei Zhu, Fei Ye, Yuchen Fu, Quan Liu, and Bairong Shen.
\newblock Electrocardiogram generation with a bidirectional lstm-cnn generative
  adversarial network.
\newblock \emph{Scientific reports}, 9\penalty0 (1):\penalty0 1--11, 2019.

\end{thebibliography}
\end{document}